\documentclass[10pt,twocolumn,letterpaper]{article}

\usepackage[pagenumbers]{cvpr} 

\usepackage{graphicx}
\usepackage{amsmath}
\usepackage{amssymb}
\usepackage{booktabs}
\usepackage{caption}
\usepackage{multirow}
\usepackage[accsupp]{axessibility}  

\usepackage[pagebackref,breaklinks,colorlinks]{hyperref}

\usepackage[capitalize]{cleveref}
\crefname{section}{Sec.}{Secs.}
\Crefname{section}{Section}{Sections}
\Crefname{table}{Table}{Tables}
\crefname{table}{Tab.}{Tabs.}

\usepackage{cuted}

\def\cvprPaperID{1119} 
\def\confName{CVPR}
\def\confYear{2023}

\begin{document}

\title{Zero-shot Generative Model Adaptation via Image-specific Prompt Learning}

\author{
  Jiayi Guo$^{1}$\thanks{Equal contribution.}\ \ \
  Chaofei Wang$^{1}$\footnotemark[1]\ \ \
  You Wu$^{2}$\ \ \
  Eric Zhang$^{3}$\ \ \
  Kai Wang$^{3}$ \ \ \
  Xingqian Xu$^{3}$ \ \ \
    Shiji Song$^{1}$ \ \ \\ 
    Humphrey Shi$^{3,4\dagger}$ \ \ \ 
  Gao Huang$^{1}$\thanks{Corresponding authors.}\\
  {\small
    $^{1}$Tsinghua University, BNRist\ \ \
    $^{2}$UCAS\ \ \
    $^{3}$SHI Labs @ Oregon \& UIUC\ \ \
    $^{4}$Picsart AI Research (PAIR)}\\
     {\small \textbf{\url{https://github.com/Picsart-AI-Research/IPL-Zero-Shot-Generative-Model-Adaptation}}}
  \vspace{-16mm}
}

\maketitle

\begin{strip}
    \centering
    \includegraphics[width=0.99\linewidth]{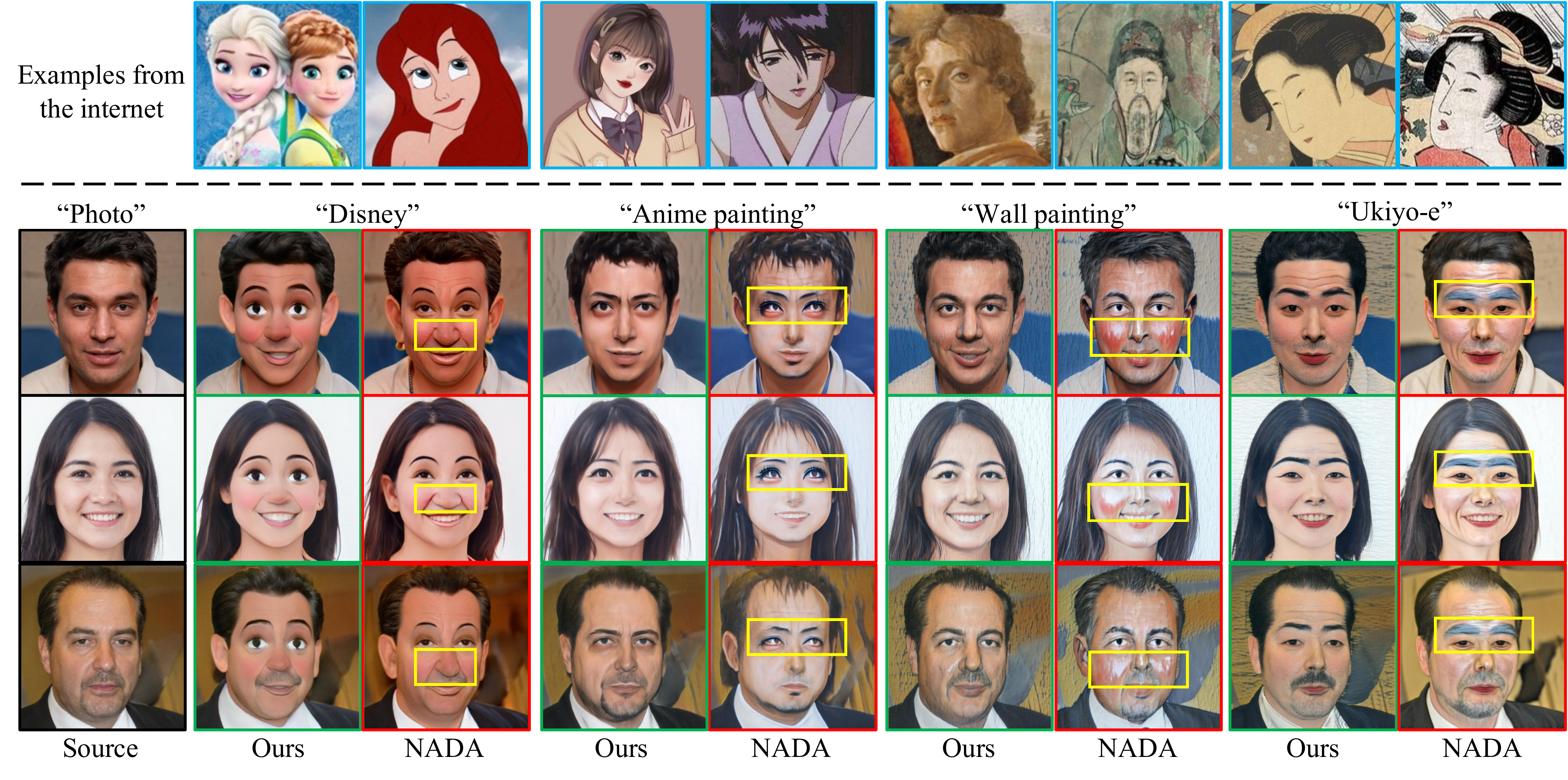}
    \vspace{-3mm}   
  \captionof{figure}{The mode collapse issue. For NADA \cite{karras2020analyzing} and our method, the same generator pre-trained on the source domain of ``Photo" is adapted to the unseen target domains of ``Disney", ``Anime painting", ``Wall painting" and ``Ukiyo-e" only with the domain labels. The images above the dotted line are some examples from the internet. The generated images of NADA exhibit some similar unseen patterns (yellow box areas) which are undesired in terms of quality and diversity. This issue is largely addressed by our method.}
  \label{fig:fig1}
\end{strip}

\begin{abstract}
Recently, CLIP-guided image synthesis has shown appealing performance on adapting a pre-trained source-domain generator to an unseen target domain. It does not require any target-domain samples but only the textual domain labels. The training is highly efficient, e.g., a few minutes. However, existing methods still have some limitations in the quality of generated images and may suffer from the mode collapse issue. 
A key reason is that a fixed adaptation direction is applied for all cross-domain image pairs, which leads to identical supervision signals. To address this issue, we propose an \textbf{I}mage-specific \textbf{P}rompt \textbf{L}earning (IPL) method, which learns specific prompt vectors for each source-domain image. This produces a more precise adaptation direction for every cross-domain image pair, endowing the target-domain generator with greatly enhanced flexibility. 
Qualitative and quantitative evaluations on various domains demonstrate that IPL effectively improves the quality and diversity of synthesized images and alleviates the mode collapse. Moreover, IPL is independent of the structure of the generative model, such as generative adversarial networks or diffusion models. 
Code is available at 
\href{https://github.com/Picsart-AI-Research/IPL-Zero-Shot-Generative-Model-Adaptation}{https://github.com/Picsart-AI-Research/IPL-Zero-Shot-Generative-Model-Adaptation}.
\end{abstract}

\section{Introduction}

In recent years, image synthesis using generative adversarial networks (GANs) \cite{goodfellow2014generative} has been rapidly developed. The state-of-the-art methods can generate images that are hard to be distinguished from real data \cite{isola2017image,karras2019style,karras2020analyzing,walton2022stylenat,xu2023image}. However, the GAN-based methods heavily rely on \textit{vast quantities} of training examples, and adopt a cumbersome adversarial training scheme which generally costs many hours of training time. Unfortunately, in many real-world scenarios, data acquisition is difficult or expensive. For example, in the artistic domains, it is impossible to have artists make thousands of creations. The high training cost is also unacceptable on some embedded devices, e.g., cellphones.

To address these issues, researchers begin to focus on the generative model adaptation. The goal of this task is to adapt a pre-trained source-domain generator to a target domain with \textit{limited} data.
 Many few-shot GAN-based methods are proposed, such as TGAN \cite{wang2018transferring}, FreezeD \cite{mo2020freeze}, MinGAN \cite{wang2020minegan}, ADA \cite{karras2020training}, DiffAug \cite{zhao2020differentiable}, IDC \cite{ojha2021few} and RSSA \cite{xiao2022few}, 
etc. However, these methods still require some training images of the target domain and follow the adversarial training scheme. 
As a pioneer work, StyleGAN-NADA \cite{gal2022stylegan} (NADA for short) proposes a \textit{zero-shot} adaptation method, which only requires textual domain labels and discards the cumbersome adversarial training scheme by introducing a pre-trained CLIP model. Although efficient, it still has obvious deficiencies, i.e., the limited quality and mode collapse of generated images. As shown in \cref{fig:fig1}, we adapt a pre-trained generator of ``Photo" domain to ``Disney", ``Anime painting", ``Wall painting" and ``Ukiyo-e" domains. For the results of NADA \cite{gal2022stylegan}, we notice that the generated images of the same target domain always show some homogeneous patterns which degrade the image quality and diversity, such as deep nasolabial folds in ``Disney", squinting eyes in ``Anime painting", red cheeks in ``Wall painting" and blue eyebrows in ``Ukiyo-e" (yellow box areas). 

\begin{figure}[t]
  \centering
  \includegraphics[width=\linewidth]{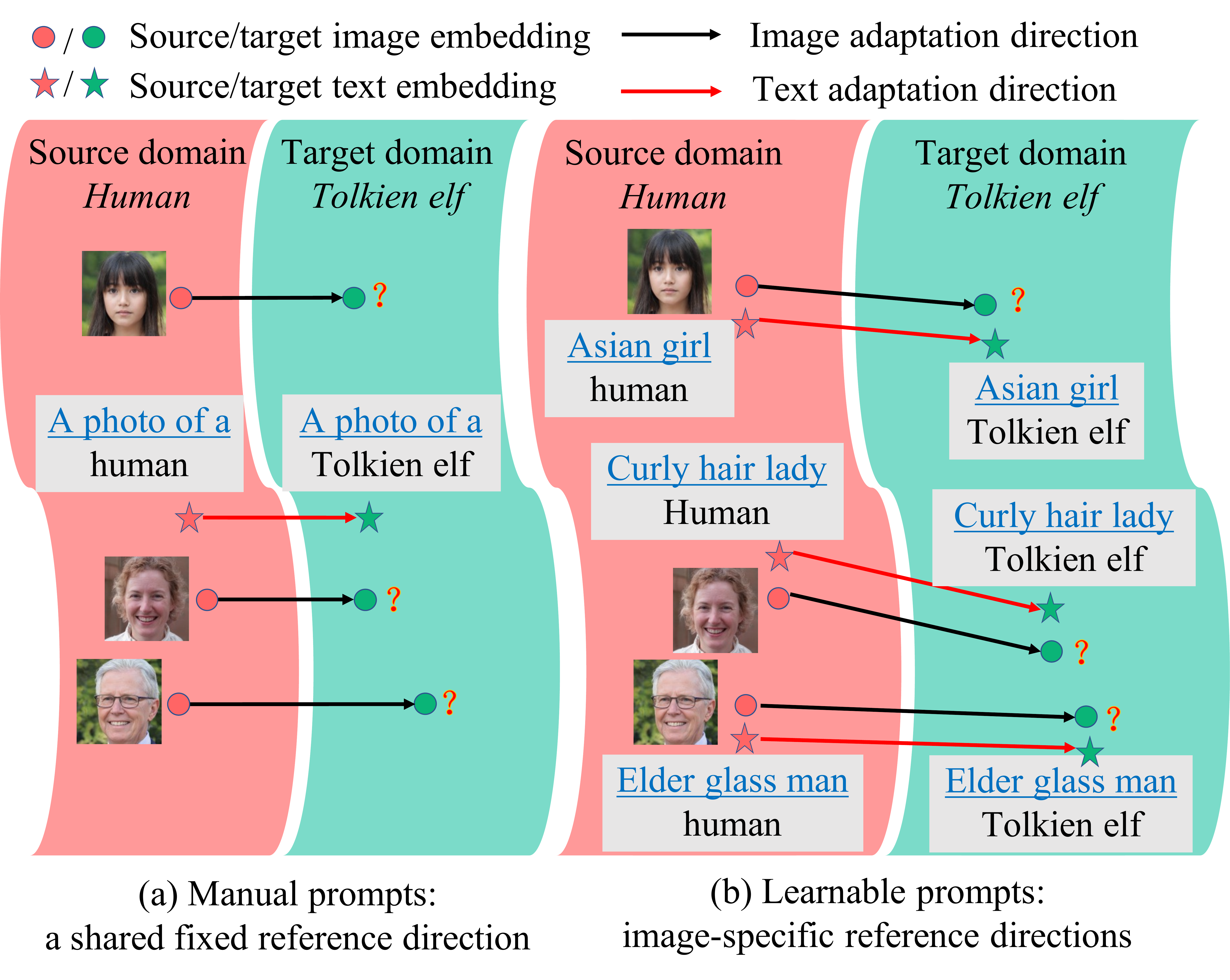}
    \vspace{-4mm}   
  \caption{An illustration of our motivation. The previous methods adopt manual prompts to compute a fixed adaptation direction for all cross-domain image pairs, while our method learns image-specific prompts for producing more precise and diversified adaptation directions.}
  \label{fig:motivation}
  \vspace{-4mm}
\end{figure}

By exploring the factors behind this phenomenon, we find that the key factor is the \textit{fixed} adaptation direction produced by manually designed prompts. Sharing the direction for all cross-domain image pairs leads to identical supervision signals for the model adaptation. 
Consider the example, adapting a generator of ``Human'' domain to ``Tolkien elf'' domain as shown in \cref{fig:motivation}. The previous works \cite{gal2022stylegan,kim2022diffusionclip} adopt manually designed prompts (e.g., ``A photo of a'') plus the domain label to produce a fixed adaptation direction, which is \textit{shared} by all cross-domain image pairs (\cref{fig:motivation} (a)) in the adaptation process. We argue that the constraint is too restrictive and suppresses the image-specific features, leading to homogeneous generated patterns.

In this paper, we propose an \textbf{I}mage-specific \textbf{P}rompt \textbf{L}earning (IPL) method to address the above issue. The motivation is setting more precise and diversified adaptation directions by customizing more image-specific prompts, for instance ``Asian girl'', ``Curly hair lady'' and ``Elder glass man'' (\cref{fig:motivation} (b)). These adaptation directions endow the target-domain generator with high flexibility to synthesize more diversified images. The proposed IPL is a two-stage method. 
In Stage 1, a latent mapper is trained to produce an \textit{image-specific} set of prompt vectors conditioned on each source-domain image by a contrastive training scheme. The learned prompt vectors contain more specific and diversified features of the source-domain images than the fixed prompt vectors. We further propose a domain regularization loss to ensure that the learned prompt vectors are compatible with the target domain. In Stage 2, we compute more \textit{precise} and \textit{diversified} adaptation directions for each cross-domain image pair, and train the target-domain generator with an adaptive directional CLIP loss, which can be viewed as an improved version of the Directional CLIP Loss \cite{gal2022stylegan}. As shown in \cref{fig:fig1}, our method alleviates the mode collapse issue well. Extensive experiments across a wide range of domains demonstrate that the proposed IPL effectively improves the quality of synthesized images and overcomes the mode collapse issue. User studies and ablation studies are also conducted to validate the effectiveness of our method.

It is worth noting that our proposed IPL method is independent of the structure of the generative model, and can be applied to the recent diffusion models \cite{sohl2015deep,song2019generative,ho2020denoising,song2020denoising,nichol2021improved,preechakul2022diffusion,xu2022versatile,liu2023more}. Thus we also combine IPL with diffusion models and get a more robust and stronger generative capacity, especially on complex images, which shows the high effectiveness and adaptability of our approach.
\section{Related Work}

\noindent\textbf{Generative model adaptation.} Generative model adaptation is the task of adapting a generative model trained on a large-scale source domain to a data-limited target domain. According to the size of the training dataset of the target domain, it can be directly divided into two main categories: few-shot generative model adaptation and zero-shot generative model adaptation. For the few-shot generative model adaptation task, the most natural approach is to fine-tune a pre-trained GAN \cite{wang2018transferring,bartunov2018few,liang2020dawson,clouatre2019figr}. However, fine-tuning the entire network weights used to result in overfitting. Subsequently, many methods were proposed to alleviate the overfitting issue. They either imposed strong regularization \cite{zhao2021improved,zhang2019consistency}, or modified the network parameters with a slight perturbation \cite{mo2020freeze,wang2020minegan,noguchi2019image,robb2020few}, or preserved some important information by cross-domain alignment \cite{ojha2021few,xiao2022few}, or performed data augmentation \cite{tran2021data,zhao2020differentiable,zhao2020image}.  
For the zero-shot generative model adaptation task, NADA \cite{gal2022stylegan} first proposed to introduce a pre-trained CLIP model for supplying necessary prior knowledge. It only required textual domain labels, and encoded the domain gap as a text-guided adaptation direction in CLIP space. 
To enhance the identity-preserving capability of real-world image translation, Kim \etal further proposed DiffusionCLIP \cite{kim2022diffusionclip} which utilized diffusion models \cite{song2020denoising} instead of StyleGANs \cite{karras2019style,karras2020analyzing,karras2020training,karras2021alias} in NADA. Nevertheless, these existing works all adopt a fixed adaptation direction which only contains the basic domain knowledge but no image-specific features. In this paper, we argue that this shared fixed adaptation direction may lead to the mode collapse issue. To produce more accurate and adaptive adaptation directions, we propose to learn diverse and specific prompt vectors for each image.

\noindent\textbf{Prompt learning.}
Prompt engineering is first introduced as a knowledge probing approach \cite{petroni2019language}. Given cloze-style prompts, it induces pre-trained language models to generate the corresponding answers. However, manually designed prompts may be sub-optimal and provide imprecise guidance. To tackle this issue, prompt learning \cite{shin2020autoprompt,jiang2020can,li2021prefix,zhong2021factual,lester2021power,gao2020making,liu2021gpt} has been widely studied in natural language processing to automatically explore the optimal set of prompts. With the unprecedented development of vision-language models \cite{radford2021learning,jia2021scaling} in recent years, researchers begin to apply prompt learning to computer vision tasks \cite{zhou2021learning,zhou2022conditional,du2022learning,li2022grounded,ju2021prompting,ge2022domain}. In specific, Zhou \etal \cite{zhou2021learning,zhou2022conditional} first adopted context optimization in image classification tasks by modeling context words with continuous vectors in the word embedding space. Subsequently, many downstream tasks in computer vision were also explored, e.g., object detection \cite{du2022learning}, visual grounding \cite{li2022grounded}, video understanding \cite{ju2021prompting} and transfer learning \cite{ge2022domain}. 
As far as we know, this is the first work to propose an adaptive prompt learning scheme for generative model adaptation. Different from previous prompt learning schemes, our method introduces a latent mapper to learn a specific set of prompt vectors for each image. When training the target-domain generator, the learned image-specific prompt vectors could produce more precise adaptation directions to provide better supervision signals.

\section{Methodology}

The goal of zero-shot generative model adaptation is to adapt a pre-trained source-domain generator $G_{\rm s}$ to an unseen target domain, and get the target-domain generator $G_{\rm t}$. The source domain with the domain label ${\rm Y}_{\rm s}$, e.g., ``Human", can obtain plentiful high-quality images by $G_{\rm s}$. The target domain is described only through the domain label ${\rm Y}_{\rm t}$, e.g., ``Tolkien elf", with no images. Following \cite{gal2022stylegan, kim2022diffusionclip}, a pre-trained CLIP model \cite{radford2021learning} including an image encoder $E_{\rm I}$ and a text encoder $E_{\rm T}$ is introduced.

\begin{figure*}[t]
  \centering
   \includegraphics[width=0.95\linewidth]{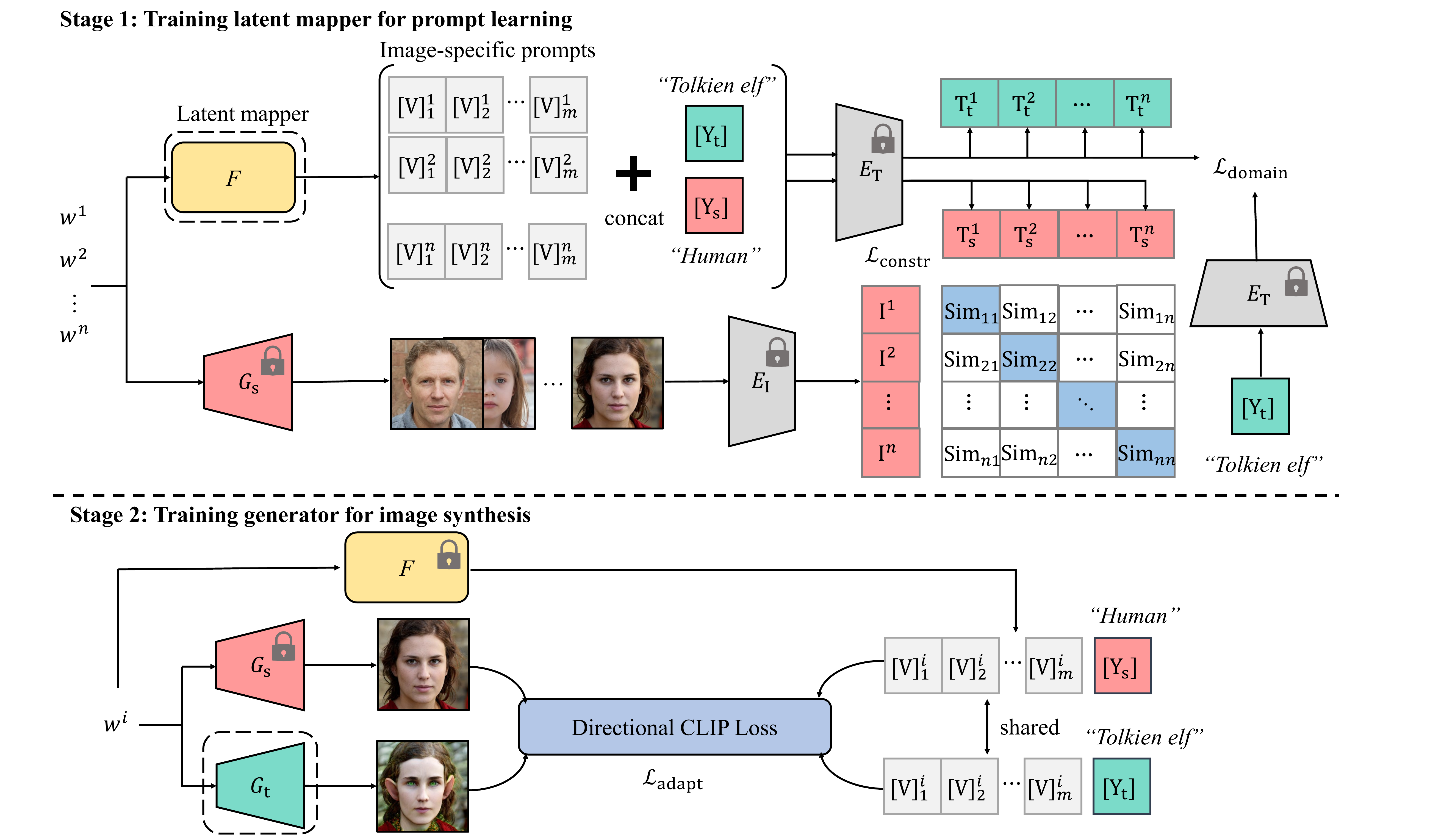}
   \caption{The framework of our method. In Stage 1, a latent mapper $F$ is trained for prompt learning by a contrastive learning loss $\mathcal{L}_{\rm contr}$ and a domain regularization loss $\mathcal{L}_{\rm domain}$. The image encoder $E_{\rm I}$ and the text encoder $E_{\rm T}$ are from the CLIP model \cite{radford2021learning}. In Stage 2, the target-domain generator $G_{\rm t}$ is trained for image synthesis by the improved Directional CLIP Loss $\mathcal{L}_{\rm adapt}$ in which the adaptive prompts produced by the latent mapper are applied. In two stages, the locked modules are fixed while the unlocked modules are trained. For simplicity, we replace $E_{\rm I}(G_{\rm s}(w^i))$ and $E_{\rm T}({\rm M}_{\rm s}^i)$ with ${\rm I}^i$ and${\rm T}_{\rm s}^i$, respectively.}
   \label{fig:method}
   \vspace{-2mm}
\end{figure*}

We propose a two-stage method named \textbf{I}mage-specific \textbf{P}rompt \textbf{L}earning (IPL). Its framework is shown in \cref{fig:method}.
In Stage 1, a latent mapper $F$ is trained to produce a set of image-specific prompt vectors $\{[\textbf{V}]_1^i,[\textbf{V}]_2^i,\cdots, [\textbf{V}]_{m}^i\}$ for each latent code $w^{i}$ of a source-domain image. Each prompt vector has the same dimension with word embeddings in CLIP space. The training loss consists of a contrastive learning loss $\mathcal{L}_{\rm contr}$ and a domain regularization loss $\mathcal{L}_{\rm domain}$. The former aims to preserve the image-specific features of each source domain image in the learned prompt vectors. The latter constrains the image-specific features to be suitable to the target domain, which means the learned features should not conflict with the target domain. For example, the features of prompts like ``round ear" should not be contained in the ideal prompt vectors if the target domain is ``Tolkien elf". In Stage 2, the trained latent mapper $F$ is plugged into the training process of the target-domain generator $G_{\rm t}$, and produces more precise and diversified adaptation directions for cross-domain image pairs. This training stage follows \cite{gal2022stylegan} except that learned prompt vectors produced by the latent mapper $F$ replace the fixed prompt vectors. The final textual supervision information includes shared learned prompt vectors and respective embeddings of the original domain labels.

\subsection{Image-specific prompt learning}
\label{sec:mapper}
\textbf{General prompts.} The previous methods \cite{gal2022stylegan,kim2022diffusionclip} compute a fixed adaptation direction produced by two embeddings of manually designed prompts, e.g., ``a photo of a human" and ``a photo of a Tolkien elf", then constrain the directions of all cross-domain pairs to be parallel with the adaptation direction. In contrast to manually designed prompts, prompt learning \cite{zhou2021learning} aims to find the optimal set of prompt vectors for a domain by directly tuning the embeddings of prompts. Formally, we define a general prompt matrix $\rm M_{\rm d}$ to represent a given domain ${\rm d}$. $\rm M_{\rm d}$ consists of the prompt vectors $[\textbf{V}]_1,[\textbf{V}]_2,\cdots, [\textbf{V}]_{m}$ and the embedding of the domain label $[{\rm Y}_{\rm d}]$ as below:
\vspace{-1ex}
\begin{equation}
    {\rm M}_{\rm d}=[\textbf{V}]_1[\textbf{V}]_2\cdots [\textbf{V}]_{m}[{\rm Y}_{\rm d}],
    \vspace{-1ex}
\label{eq:Md}
\end{equation}
where $m$ is the number of prompts. Suppose the dimension of each embedding is $k$. Then the dimension of ${\rm M}_{\rm d}$ should be $(m+1)\times k$. In \cite{gal2022stylegan,kim2022diffusionclip}, the prompt vectors $[\textbf{V}]_1,[\textbf{V}]_2,\cdots, [\textbf{V}]_{m}$ are fixed embeddings of manually designed prompts. For prompt learning \cite{zhou2021learning}, the prompt vectors are learned by encoding each training image of the domain ${\rm d}$ with $E_ {\rm I}$ and the prompt matrix $\rm M_{\rm d}$ with $E_ {\rm T}$, and then maximizing the cosine similarity between them.

Inspired by prompt learning, in the zero-shot generative model adaptation task, a natural idea is to learn an optimal set of prompt vectors instead of the manually designed prompts in NADA \cite{gal2022stylegan}. Although the adaptation direction calculated by the learned prompt vectors seems to be more reasonable than that of the manually designed prompts, it is still fixed and shared for all cross-domain image pairs. These fixed learned prompt vectors can not solve the mode collapse issue (Experimental validations can be seen in \cref{sec:ablation}). To obtain more flexible and diversified adaptation directions, we further propose to learn a set of image-specific prompt vectors for each image, which can be regarded as an improved version of prompt learning.

\textbf{Image-specific prompts.} Utilizing the source-domain generator $G_{\rm s}$, we train a latent mapper $F$ as shown in \cref{fig:method} (Stage 1). Through the mapper, each image of the source domain can be matched to an optimal set of prompt vectors. Formally, given a latent code $w^i$, corresponding to the $i^{\rm th}$ image in the source domain, the image-specific set of prompt vectors $\{[\textbf{V}]_1^i,[\textbf{V}]_2^i,\cdots, [\textbf{V}]_{m}^i\}$ can be obtained by $F(w^i,\theta)$, where $\theta$ denotes the parameters of the latent mapper $F$. Following the definition of the prompt matrix in \cref{eq:Md}, we define an image-specific prompt matrix of the $i^{\rm th}$ source-domain image as:
\vspace{-1ex}
\begin{equation}
    {\rm M}_{\rm s}^{i} = F(w^i,\theta)[{\rm Y}_{\rm s}] = [\textbf{V}]_1^i[\textbf{V}]_2^i\cdots [\textbf{V}]_{m}^i[{\rm Y}_{\rm s}].
\label{equ:prompt}
\vspace{-1ex}
\end{equation}
In this paper, $F$ is a common four-layer fully-connected network. Next, we show how to train it.

\textbf{Contrastive training scheme.} Given a batch of latent codes $\{w^1,w^2,...,w^n\}$, we can produce a batch of sets of prompt matrices $\{{\rm M}^1_{\rm s},{\rm M}^2_{\rm s},...,{\rm M}^n_{\rm s}\}$ by $F$ and a batch of images $\{G_{\rm s}(w^1),G_{\rm s}(w^2),...,G_{\rm s}(w^n)\}$ by $G_{\rm s}$. Then $n\times n$ pairs $<G_{\rm s}(w^i), {\rm M}^j_{\rm s}>$, $i,j \in \{1,2,...,n\}$ have been obtained. Then, we take the pairs of $i=j$ as positive samples, and the pairs of $i\neq j$ as negative samples for contrastive training. Specifically, we compute the similarity between embeddings of the $i^{\rm th}$ image and the $j^{\rm th}$ prompt matrix in CLIP space as:
 \vspace{-1ex}
\begin{equation}
\small
    {\rm Sim}_{ij} = {\rm Cos}({\rm Norm} (E_{\rm I}(G_{\rm s}(w^i))), {\rm Norm}(E_{\rm T}({\rm M}^j_{\rm s}))),
    \vspace{-1ex}
\end{equation}
where ${\rm Norm}(\cdot)$ and ${\rm Cos}(\cdot)$ represent $L_2$ normalization and the cosine function, respectively. 
The similarities of positive samples are maximized while the similarities of negative samples are minimized. The contrastive loss is expressed as:
\vspace{-1ex}
\begin{equation}
    \mathcal{L}_{\rm contr} = \mathbb{E}_{w \in \mathcal{W}}(\sum_{i\neq j}({\rm Sim}_{ij})-\sum_{i=j}({\rm Sim}_{ij})).
    \vspace{-1ex}
\end{equation}

\textbf{Domain regularization loss.} For the target domain without any prior knowledge except the domain label ${\rm Y}_{\rm t}$, we can simply share the learned prompt vectors between the source and target domains following \cite{gal2022stylegan}. However, the shared prompt vectors may lead to the risk of generating unrealistic images for the target domain, because some learned prompt vectors may contain strongly relevant features to the source domain, leading to conflict with the target domain. For example, an image of ``Human" domain is matched to prompt vectors of ``round ear", but a corresponding image of ``Tolkien elf" domain should not contain the features of ``round ear". Sharing these prompt vectors is harmful to the target-domain image generation. Therefore, we further propose a domain regularization loss. Specifically, we constrain the angles between the embeddings of the image-specific prompt matrix ${\rm M}_{\rm t}^i$ and the target-domain label ${\rm Y}_{\rm t}$ in CLIP space to be small, to avoid the learned prompt vectors conflicting with the target domain. Formally, the domain regularization loss is described as: 
\begin{equation}\label{equ:domain}
    \mathcal{L}_{\rm domain}=-\mathbb{E}_{w^i \in \mathcal{W}}\sum_{i=1}^{n}( {\rm Cos}(E_{\rm T}({\rm M}_{\rm t}^i),E_{\rm T}({\rm Y}_{\rm t}))),
\end{equation}
where ${\rm M}_{\rm t}^i$ is calculated by \cref{equ:prompt} except replacing the domain label, ${\rm Cos}(\cdot)$ represents the cosine similarity.

As a summary, the whole training loss function of the latent mapper $F$ is:
\begin{equation}\label{equ:sum}
    \mathcal{L} = \mathcal{L}_{\rm constr} + \lambda\mathcal{L}_{\rm domain},
\end{equation}
where $\lambda$ is the ratio parameter. Optimized by $\mathcal{L}$, the learned prompt vectors can not only reflect the features of the source-domain images, but also adapt to the target domain.

\subsection{Latent mapper guided generator training}
\label{sec:stage2}
After training the latent mapper $F$, we conduct the second stage: training the target-domain generator $G_{\rm t}$ as shown in \cref{fig:method} (Stage 2). In specific, we plug in the trained latent mapper, and train $G_{\rm t}$ with an improved Directional CLIP Loss $\mathcal{L}_{\rm adapt}$. Its main difference with \cite{gal2022stylegan} is using the image-specific prompt vectors that are produced on-the-fly by $F$ instead of the fixed ones of manually designed prompts. 
Formally, given a latent code $w^i$, we calculate the direction of the $i^{\rm th}$ source and target image pair as below:
\begin{equation}
    \Delta {\rm I}_{i} = {\rm Norm} (E_{\rm I}(G_{\rm t}(w^i))-{\rm Norm}(E_{\rm I}(G_{\rm s}(w^i)),
\end{equation}
where ${\rm Norm}(\cdot)$ represents $L_2$ normalization. The image-specific adaptation direction is calculated as below:
\begin{equation}
    \Delta {\rm T}_{i} = {\rm Norm} (E_{\rm T}({\rm M}_{\rm t}^{i}))-{\rm Norm}(E_{\rm T}({\rm M}_{\rm s}^{i})).
\end{equation}
The improved Directional CLIP Loss $\mathcal{L}_{\rm adapt}$ is:
\begin{equation}
    \mathcal{L}_{\rm adapt}=\mathbb{E}_{w^i \in \mathcal{W}}\sum_{i=1}^{n} (1-\frac{\Delta {\rm I}_{i} \cdot \Delta {\rm T}_{i}}{|\Delta {\rm I}_{i}||\Delta {\rm T}_{i}|}),
\end{equation}
where $n$ is the batch size of latent codes. $\mathcal{L}_{\rm adapt}$ constrains the direction of each image pair $\Delta {\rm I}_{i}$ with an image-specific adaptation direction $\Delta {\rm T}_{i}$. 

\begin{figure*}[t]
  \centering
  \includegraphics[width=0.98\linewidth]{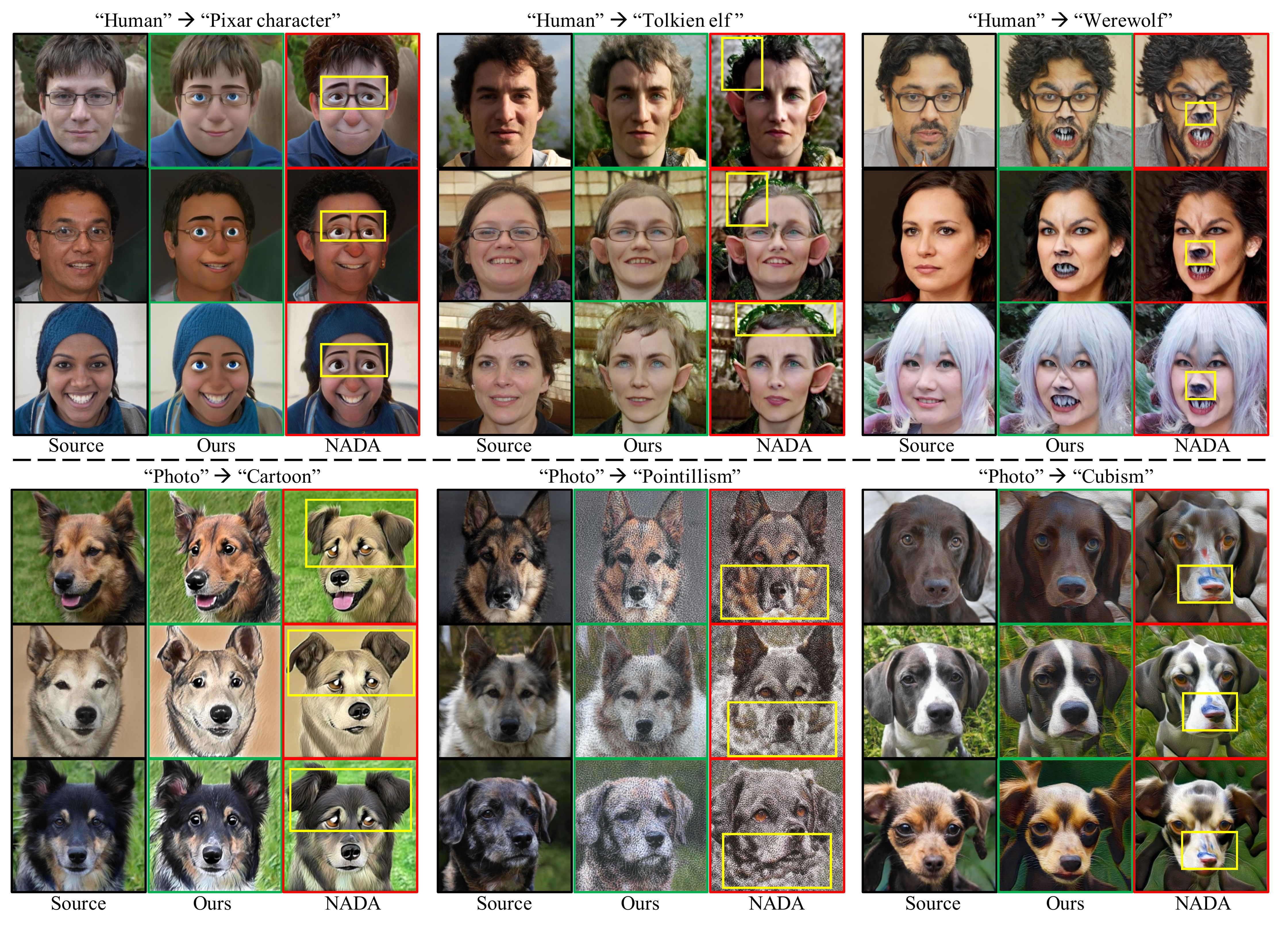}
    \vspace{-4mm}   
  \caption{Image synthesis comparison results. For FFHQ \cite{karras2020analyzing}, the source domain is ``Human" and the target domains are ``Pixar character", ``Tolkien elf", and ``Werewolf". For AFHQ-Dog \cite{choi2020stargan}, the source domain is ``Photo" and the target domains are ``Cartoon", ``Pointillism", and ``Cubism". The yellow box areas show the mode collapse problem of NADA \cite{gal2022stylegan}.}
  \label{fig:others}
  \vspace{-2mm}
\end{figure*}

\begin{table*}[t]
\begin{center}
\scriptsize
\caption{Quantitative evaluation results. US denotes user study. The best results are \textbf{bold}.}
  \vspace{-2mm}
\label{table:qer}
\renewcommand{\arraystretch}{1.2}

\begin{tabular}{ccccccccccccccc}
\toprule

  \multirow{3}*{Dataset}&\multirow{3}*{Source$\rightarrow$Target}&\multicolumn{2}{c}{{IS \cite{salimans2016improved} ($\uparrow$)}}&\multicolumn{2}{c}{{SCS \cite{xiao2022few} ($\uparrow$)}}&\multicolumn{2}{c}{{ID \cite{deng2019arcface,he2022transfg} ($\uparrow$)}}&\multicolumn{6}{c}{{SIFID  \cite{shaham2019singan} ($\downarrow$)}}&\multirow{3}*{US ($\uparrow$)}\\
  \cline{3-14}
  
  &&\multirow{2}{*}{NADA}&\multirow{2}{*}{IPL}&\multirow{2}{*}{NADA}&\multirow{2}{*}{IPL}&\multirow{2}{*}{NADA}&\multirow{2}{*}{IPL}&\multicolumn{3}{c}{NADA}&\multicolumn{3}{c}{IPL}\\
  \cline{9-14}
  &&&&&&&&$\rm{R_1}$&$\rm{R_2}$&$\rm{R_3}$&$\rm{R_1}$&$\rm{R_2}$&$\rm{R_3}$\\
 \hline
 \multirow{7}{*}{FFHQ \cite{gal2022stylegan}}
 &Photo$\rightarrow$Disney&2.721&\textbf{3.089}&0.407&\textbf{0.448}&0.782&\textbf{0.801}&2.776&3.136&3.670&\textbf{2.517}&\textbf{2.930}&\textbf{3.497}&82.6\%\\
&Photo$\rightarrow$Anime painting&2.450&\textbf{3.051}&0.324&\textbf{0.518}&0.666&\textbf{0.776}&2.956&1.811&1.242&\textbf{2.845}&\textbf{1.595}&\textbf{1.021}&79.3\%\\
&Photo$\rightarrow$Wall painting&2.183&\textbf{2.676}&0.439&\textbf{0.487}&0.594&\textbf{0.637}&1.944&1.220&1.331&\textbf{1.930}&\textbf{1.183}&\textbf{1.274}&80.9\%\\
&Photo$\rightarrow$Ukiyo-e&2.205&\textbf{2.974}&0.420&\textbf{0.506}&\textbf{0.775}&0.632&1.954&1.990&1.326&\textbf{1.165}&\textbf{1.255}&\textbf{0.878}&85.9\%\\

&Human$\rightarrow$Pixar character&2.703&\textbf{2.785}&0.379&\textbf{0.461}&0.757&\textbf{0.853}&0.793&0.932&\textbf{0.865}&\textbf{0.638}&\textbf{0.821}&1.092&86.7\%\\
&Human$\rightarrow$Tolkien elf&2.479&\textbf{2.778}&0.416&\textbf{0.491}&0.711&\textbf{0.772}&\textbf{0.632}&1.495&1.452&0.690&\textbf{0.637}&\textbf{0.701}&76.8\%\\

&Human$\rightarrow$Werewolf&2.619&\textbf{2.809}&0.399&\textbf{0.417}&0.642&\textbf{0.747}&1.969&1.846&1.967&\textbf{1.734}&\textbf{1.688}&\textbf{1.911}&72.7\%\\
\hline
\multirow{3}{*}{AFHQ \cite{choi2020stargan}}&Photo$\rightarrow$Cartoon&6.505&\textbf{8.658}&0.407&\textbf{0.563}&0.925&\textbf{0.941}&2.708&2.672&3.870&\textbf{2.517}&\textbf{2.477}&\textbf{3.278}&87.6\%\\

&Photo$\rightarrow$Pointillism&5.419&\textbf{6.913}&0.224&\textbf{0.542}&0.775&\textbf{0.881}&7.081&5.288&7.142&\textbf{4.818}&\textbf{3.089}&\textbf{4.074}&78.5\%\\

&Photo$\rightarrow$Cubism&4.165&\textbf{6.450}&0.386&\textbf{0.463}&0.934&\textbf{0.943}&2.779&\textbf{2.938}&3.199&\textbf{2.431}&2.956&\textbf{2.284}&74.3\%\\

    \bottomrule
\end{tabular}
\end{center}
  \vspace{-7mm}
\end{table*}

\section{Experiments}\label{sec:exp}
In this section, we evaluate our method qualitatively and quantitatively. The experimental setup is firstly presented in \cref{sec:setup}. Then we show image synthesis results across various domains in \cref{sec:main_result}. Utilizing a GAN inversion model and diffusion models, results of real-world image translation are provided in \cref{sec:inversion}. Finally, we carefully conduct ablation studies on prompt designing schemes and loss term ratios in \cref{sec:ablation}.

\begin{figure*}[t]
  \centering
  \includegraphics[width=0.98\linewidth]{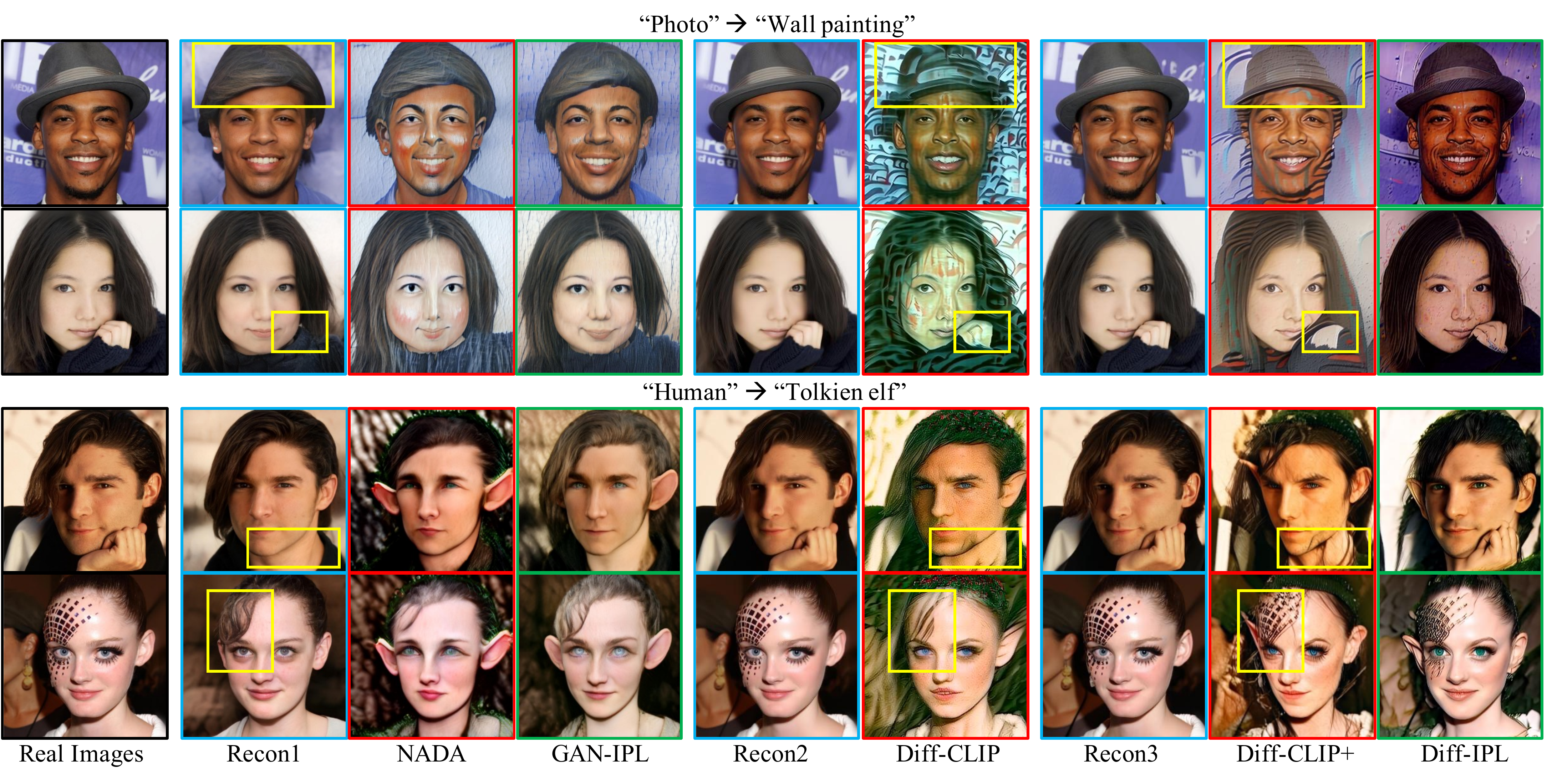}
    \vspace{-4mm}   
  \caption{Real-world image translation comparison results. Baselines are NADA \cite{gal2022stylegan}, Diff-CLIP \cite{kim2022diffusionclip} and Diff-CLIP+ (an improved version of Diff-CLIP). Recon1, Recon2 and Recon3 refer to inversion results via Restyle \cite{alaluf2021restyle}, DDIM and diffusion autoencoders, respectively. GAN-IPL and Diff-IPL denote integrating IPL with NADA and Diff-CLIP+, respectively. Real images are from CelebA-HQ dataset \cite{liu2015faceattributes} and translated into two styles of images, ``Wall painting" and ``Tolkien elf". The yellow boxes show the key observation areas.}
  \label{fig:i2i}
  \vspace{-5mm}
\end{figure*}

\subsection{Experimental setup}
\label{sec:setup}

\textbf{Baselines and settings.} Two strong methods are chosen as our competitors. For zero-shot image synthesis, NADA \cite{gal2022stylegan} is the state-of-the-art method. Following NADA \cite{gal2022stylegan}, we adapt the pre-trained StyleGANv2 \cite{karras2020analyzing} generators on (i) Flickr-Faces-HQ (FFHQ) \cite{gal2022stylegan} and (ii) Animal FacesHQ (AFHQ) \cite{choi2020stargan}, utilize the same pre-trained CLIP \cite{radford2021learning} built on ViT-B/32 \cite{dosovitskiy2020image}. For zero-shot real-world image translation, we utilize Restyle \cite{alaluf2021restyle} with e4e \cite{tov2021designing} encoder to invert a real image into the latent space $\mathcal{W}$ for StyleGANs. DiffusionCLIP (Diff-CLIP for short) \cite{kim2022diffusionclip} is the state-of-the-art method. We follow the setting of \cite{kim2022diffusionclip} except replacing denoising diffusion implicit models (DDIM) \cite{song2020denoising} with diffusion autoencoders \cite{preechakul2022diffusion}. The training process includes 300 iterations for prompt learning and 300 iterations for generator adaptation using a single NVIDIA RTX 3090 GPU. The batch size is set to 32 for prompt learning and 2 for generator adaptation. The number of learned prompt vectors $m$ is set to 4. For each domain, the ratio parameter $\lambda$ in \cref{equ:sum} is selected among [1, 10], according to the best Inception Score \cite{salimans2016improved} of adapted generators. The whole training process requires about 10$\sim$20 minutes. More implementation details can be seen in supplementary materials.

\textbf{Evaluation metrics.} The ideal generated images should have: 1) high quality and diversity, 2) correct target-domain style, and 3) necessary source-domain information preservation (e.g., structure or identity). For a comprehensive evaluation, we utilize the popular Inception Score (IS) \cite{salimans2016improved} to evaluate the image quality and diversity, the Single Image Fréchet Inception Distance (SIFID) \cite{shaham2019singan} to evaluate the target-domain style, the Structural Consistency Score (SCS) \cite{xiao2022few} to evaluate the structure preservation, the identity similarity (ID) \cite{deng2019arcface,he2022transfg} to evaluate the identity preservation. More details can be seen in supplementary materials.

\subsection{Generative model adaptation}\label{sec:main_result}

\textbf{Qualitative comparison.} In addition to \cref{fig:fig1}, we conduct extensive experiments across a wide range of domains as shown in \cref{fig:others}. All results indicate that our proposed approach outperforms NADA consistently. 
The yellow box areas in the figures denote the main different features between NADA and our IPL. From the quality of the generated images, the results of NADA have more incorrect features and noise, such as green mussy noise on hairs (Tolkien elf), ruined noses (Werewolf) and unshaped necks (Pointillism), while the results of IPL are more clear and correct. From the mode collapse perspective of the generated images, NADA is prone to collapse to some similar facial features for different images, such as depressed emotions (Pixar character), folded ears (Cartoon) and blue noses (Cubism), while IPL presents consistently higher diversity and solve the mode collapse issue well. Our advantages mainly come from the fact that the latent mapper preserves sufficient image-specific and target-domain friendly features from the source-domain images. The produced prompt vectors provide more precise and diversified adaptation directions for the target-domain generator adaptation. 

\textbf{Quantitative comparison.} To quantify the performance improvement of IPL compared to NADA \cite{gal2022stylegan}, IS, SCS, ID and SIFID are evaluated. As reported in \cref{table:qer}, for IS, IPL outperforms NADA on all 10 settings, indicating our method achieves better image quality and diversity. For SCS and ID, IPL outperforms NADA on most of the 10 settings except ``Human $\rightarrow$ Ukiyo-e''. It is mainly because that ``Ukiyo-e" naturally favors humans with narrow eyes and pale skin, which encourages identity changes during training. For SIFID, we collect 3 reference images ($\rm R_1$, $\rm R_2$, and $\rm R_3$) on the internet for each target domain. \cref{table:qer} shows that IPL outperforms NADA in most cases, indicating our superiority in generating precise target-domain styles.

\textbf{User studies.} For each target domain, 32 images generated by NADA and our method are provided to human observers, together with their corresponding source images and textual labels of target domains. Human observers are required to choose better synthesized images which are semantically more consistent with the target domain labels and preserve the useful source-domain information better. We collect 1210 responses from 121 people using a survey platform. As reported in the last column of \cref{table:qer}, 80.5$\%$ of users prefer our approach to NADA on average.

\subsection{Real-world image translation}\label{sec:inversion}

This task first inverts a real-world image to the latent code by a pre-trained inversion model and then feeds it to the trained target-domain generator to get the translated target-domain image. For GAN-based generators, we compare our method (GAN-IPL) with NADA by connecting the inversion model Restyle \cite{alaluf2021restyle}. For diffusion model generators, we compare our method (Diff-IPL) with Diff-CLIP \cite{kim2022diffusionclip} and Diff-CLIP+ which is an improved version of Diff-CLIP \cite{kim2022diffusionclip} by replacing the original DDIM \cite{song2020denoising} with a diffusion autoencoder \cite{preechakul2022diffusion}. For these diffusion models, a deterministic inversion process is naturally provided. 

As shown in \cref{fig:i2i}, comparing the results of NADA and GAN-IPL, IPL's superiority of alleviating mode collapse over NADA can still be observed. Comparing the results of Recon1, Recon2 and Recon3, diffusion models (Recon2 and Recon3) consistently perform better identity preservation than Restyle (Recon1) for real image inversion, especially for some uncommon stuffs in a human face photo, e.g., the hats, hands and tattoos in \cref{fig:i2i}. However, this property is not well inherited in the target domain generators with a fixed adaptation direction (see the results of Diff-CLIP and Diff-CLIP+). Our proposed IPL could help preserve the details in source images better and present the target-domain styles correctly (see the results of Diff-IPL). Quantitative evaluation results of Diff-CLIP, Diff-CLIP+ and Diff-IPL can be seen in supplementary materials.

\subsection{Ablation studies}\label{sec:ablation}

\begin{figure}[t]
  \centering
  \includegraphics[width=0.95\linewidth]{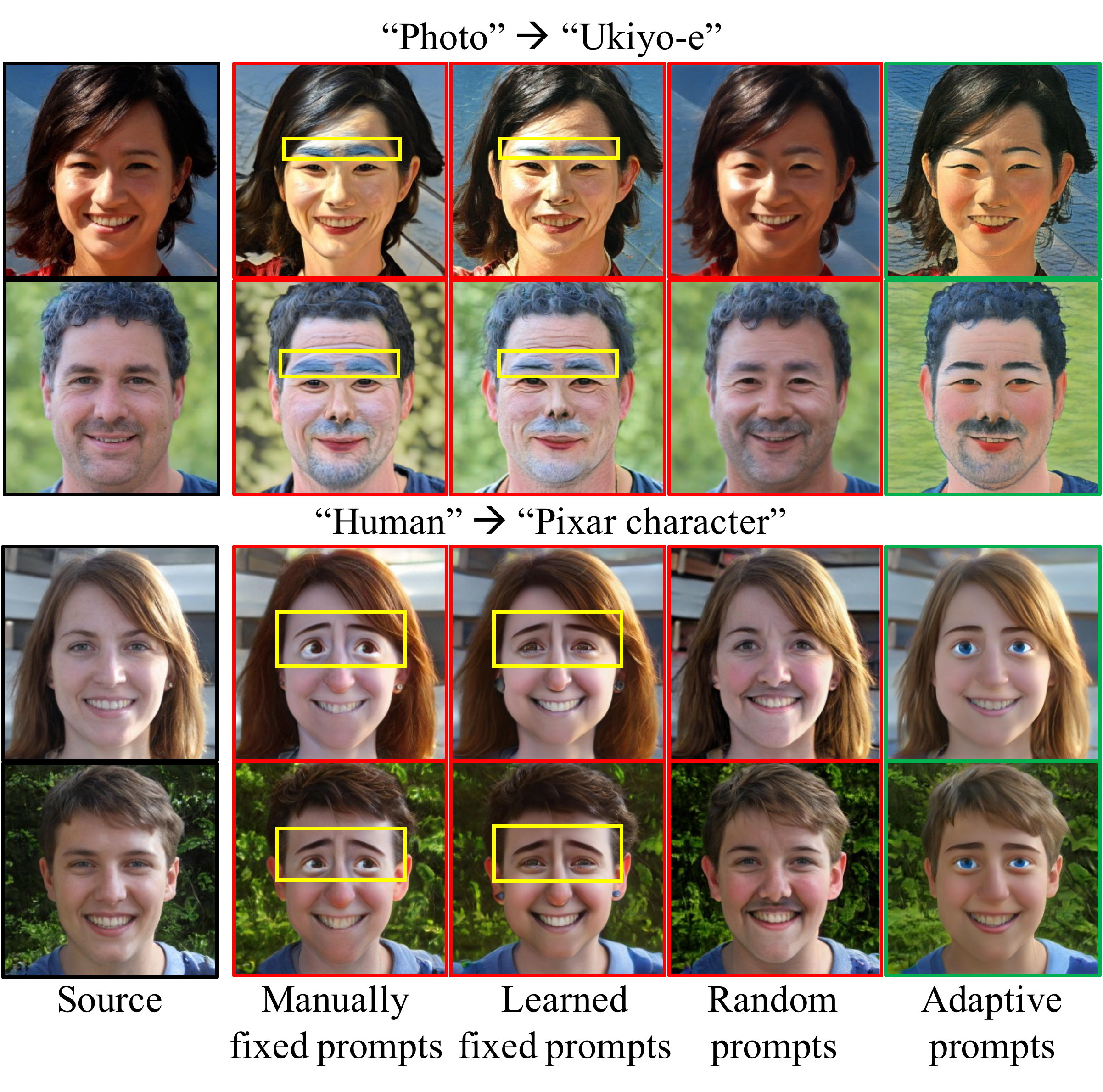}
    \vspace{-4mm}   
  \caption{Ablation results of prompt designing schemes.}
  \label{fig:abl1}
  \vspace{-4mm}
\end{figure}

\textbf{Prompt designing schemes.} We investigate four different prompt designing schemes: 1) manually fixed prompts (NADA), 2) learned fixed prompts, 3) random prompts and 4) adaptive prompts (Ours). Manually fixed prompts mean simply utilizing the manually designed prompts as NADA \cite{gal2022stylegan}. Learned fixed prompts denote unified prompt vectors produced by common prompt learning strategy \cite{zhou2021learning} and shared for all images. Random prompts refer to prompt vectors produced by a randomly initialized latent mapper. Adaptive prompts denote the learned image-specific prompt vectors produced by our IPL method.

As illustrated in \cref{fig:abl1}, synthesized images with manually fixed prompts and learned fixed prompts show some similar mode collapse issues, e.g., blue eyebrows (Ukiyo-e) and depressed emotions (Pixar character). They both produce a fixed adaptation direction, which leads to identical supervision signals for all image pairs. Synthesized images with random prompts present more photo-realistic results but lack the desired target-domain style. A possible reason is that the random prompts contain some features conflicting with the target domain and impede the learning of the target domain style. Our adaptive prompts perform best since the prompts contain more image-specific and target-domain friendly features from the source-domain images. 
  
\begin{figure}[t]
  \centering
  \includegraphics[width=0.95\linewidth]{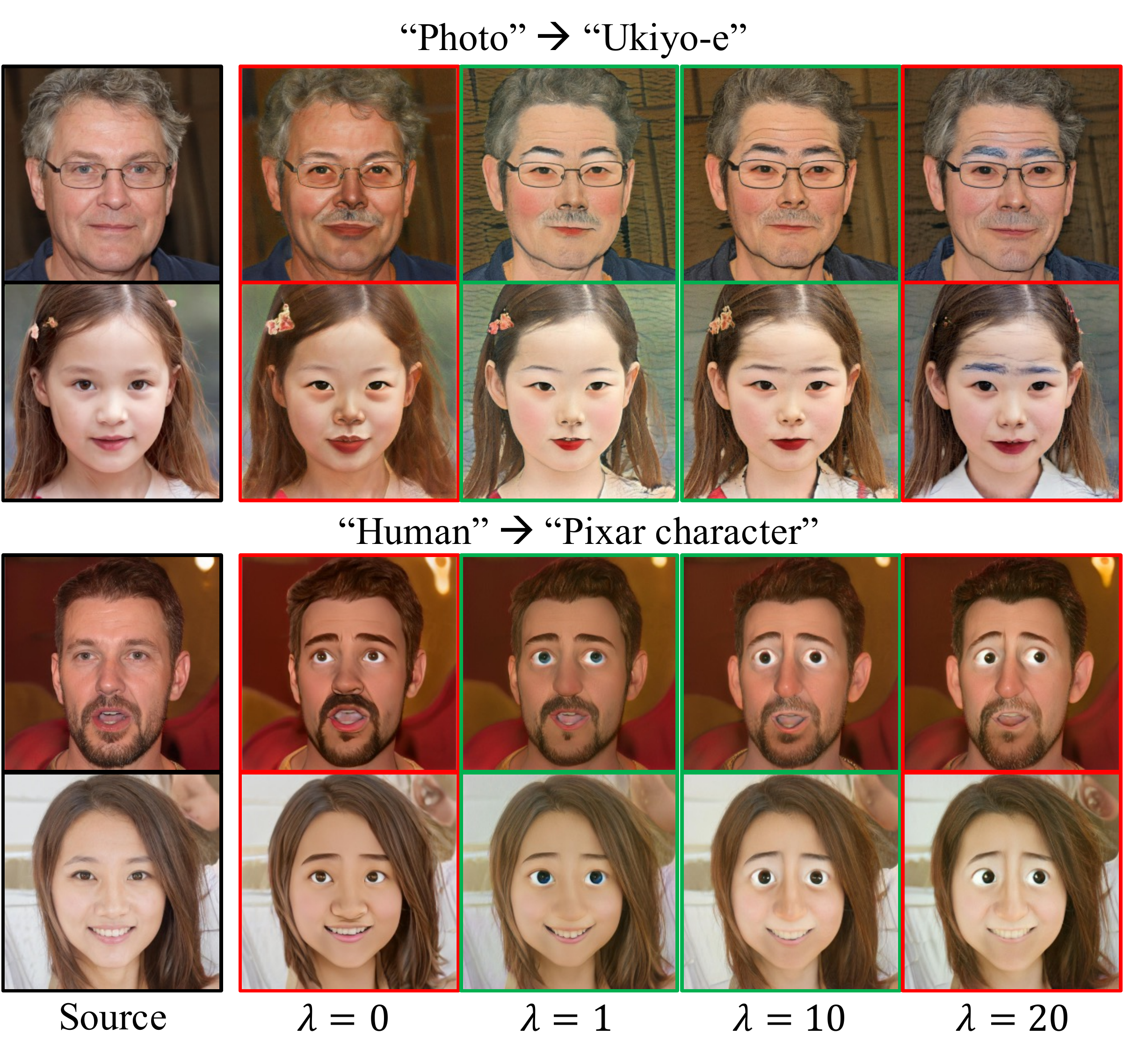}  
  \caption{Ablation results of loss term ratios.}
  \label{fig:abl2}
  \vspace{-4mm}
\end{figure}

\textbf{Loss term ratios.} We compare different values of the ratio parameter $\lambda$ in \cref{equ:sum}, which is used to adjust the intensity of the domain regularization loss. Visual results are shown in \cref{fig:abl2}. In specific, when we set $\lambda$ to a small value ($\lambda=0$ as an extreme case), there is almost no constraint from the target domain. The learned prompts would excessively preserve the source-domain features. Thus the synthesized images are similar to their corresponding source images. In contrast, if $\lambda$ is set to a large value ($\lambda=20$ as an example), a strong target-domain constraint will limit the diversity of the learned prompts. As a result, the synthesized images would slightly show some similar undesired patterns as images generated via fixed prompts. Therefore, in practical applications, $\lambda$ should be a trade-off value (i.e., between 1 and 10).

\vspace{-1mm}
\section{Conclusion}
\vspace{-1mm}
In this paper, we have proposed a novel zero-shot generative model adaptation approach called \textbf{I}mage-specific \textbf{P}rompt \textbf{L}earning (IPL). In specific, we build a projection from latent codes to image-specific sets of prompt vectors via a latent mapper. With a contrastive learning scheme and a domain regularization constraint, the learned prompt vectors represent image-specific but target-domain-friendly features, producing more precise and diversified adaptation directions for target domain generator training. Compared with the state-of-the-art approaches, IPL consistently improves the quality of synthesized images and alleviates the mode collapse issue. Furthermore, IPL is independent of the type of generator and works well with both GANs and diffusion models, which exhibits good universality and adaptability. In the future, we will try to apply the proposed image-specific prompt learning strategy in other downstream tasks, such as unsupervised image captioning. 
\vspace{-1mm}
\section*{Acknowledgements}
\vspace{-1mm}

This work is supported in part by the National Key R\&D Program of China (2019YFC1408703), the National Natural Science Foundation of China (62022048, 62276150), Guoqiang Institute of Tsinghua University and Beijing Academy of Artificial Intelligence.

{\small
\bibliographystyle{ieee_fullname}
\bibliography{egbib}
}
\clearpage
\appendix
\section*{Supplementary Materials}

\section{Detailed Experiment Settings}

We introduce the implementation details of GAN-IPL, Diff-IPL, and the evaluation metrics.

\textbf{GAN-IPL.} Our method is developed in PyTorch \cite{paszke2019pytorch}. We use the Adam \cite{kingma2014adam} optimizer with a learning rate of 0.05 for latent mapper and 0.002 for StyleGANs. The training process includes 300 iterations for prompt learning and 300 iterations for generator adaptation, using a single NVIDIA RTX 3090 GPU. The batch size is set to 32 for prompt learning and 2 for generator adaptation. The number of learned prompt vectors $m$ is set to 4. The 4 prompt vectors are initialized as the word embeddings of ``a photo of a". We use the same Layer-Freezing technique as NADA \cite{gal2022stylegan} to select the suitable training layers for each iteration and set the exponential moving average (EMA) decay \cite{tarvainen2017mean} to 0.99.
In the domain regularization loss, following CLIP \cite{radford2021learning}, we separately concatenate 79 manually designed sets of prompts (e.g., ``a photo of a ...", ``a drawing of a ...") with a domain label and feed them into $E_{\rm T}$. The average vector of the 79 encoded feature vectors replaces the encoded feature vector of the domain label $E_{\rm T}({\rm Y}_{\rm s})$ or $E_{\rm T}({\rm Y}_{\rm t})$. For each domain, the ratio parameter $\lambda$ of the domain regularization loss is selected among [1, 10], according to the best Inception Score \cite{salimans2016improved} of adapted generators. The values of $\lambda$ on different settings are provided in \cref{table:ltr}. Compared with NADA, the additional training time from the latent mapper is about 10 minutes, which is easily acceptable.

\textbf{Diff-IPL.} Applied with the Adam \cite{kingma2014adam} optimizer, the learning rates for latent mapper and diffusion autoencoders \cite{preechakul2022diffusion} are set to $\rm 7e^{-2}$ and $\rm 3e^{-5}$, respectively. The training process requires higher memory cost, utilizing a single NVIDIA A6000 GPU. Following Diff-CLIP \cite{kim2022diffusionclip}, the batch size is set to 1 for generator adaptation. We also precompute the latent codes of 50 training images via the reverse process of diffusion autoencoders and train target-domain generators for 5 epochs as \cite{kim2022diffusionclip}. We can further accelerate training with fewer diffusion discretization steps \cite{song2020denoising}. In our experiments, the number of forward steps and reverse steps are reduced to 100 and 250, respectively.

\textbf{Metrics.} We utilize Inception Score (IS) \cite{salimans2016improved}, Single Image Fréchet Inception Distance (SIFID) \cite{shaham2019singan}, Structural Consistency Score (SCS) \cite{xiao2022few} and identity similarity (ID) \cite{deng2019arcface,he2022transfg} for quantitative evaluation. In specific, for ID, we compute the identity similarity in ArcFace \cite{deng2019arcface} for FFHQ (human faces). For AFHQ (dog faces), we apply TransFG \cite{he2022transfg}, a fine-grained species recognition approach to extract identity features and compute the cosine similarity between source and target (generated) images. For SIFID, we manually collect several reference images of each target domain from the internet and compute the SIFID score for each reference image. We enclose these reference images in the folder ``reference". Although the variance of different reference images may lead to an imprecise score in some extreme cases, the superiority of an effective method could still be verified if it outperforms others in most cases.

\begin{table}[t]
\begin{center}
 \small
\caption{Loss term ratio $\lambda$ on different settings.}
\label{table:ltr}
\renewcommand{\arraystretch}{1.1}
\begin{tabular}{ccc}
\toprule
 Setting &Source$\rightarrow$Target & $\lambda$   \\
 \midrule
\multirow{10}*{GAN-IPL}& Photo$\rightarrow$Disney  &  1  \\
& Photo$\rightarrow$Anime painting & 1   \\
& Photo$\rightarrow$Wall painting &  1  \\
& Photo$\rightarrow$Ukiyo-e & 1   \\

& Human$\rightarrow$Pixar character &  1  \\
& Human$\rightarrow$Tolkien elf &  5  \\
& Human$\rightarrow$Werewolf &  5  \\
\cmidrule{2-3}
& Photo$\rightarrow$Cartoon &  10  \\
& Photo$\rightarrow$Pointillism &  10  \\
& Photo$\rightarrow$Cubism &  10  \\
\midrule
\multirow{2}*{Diff-IPL}&Photo$\rightarrow$Wall painting  & 3   \\
& Human$\rightarrow$Tolkien elf  &  2  \\
\bottomrule
\end{tabular}
\vspace{-8mm}
\end{center}
\end{table}
\begin{figure*}[ht]
  \centering
  \includegraphics[width=0.99\linewidth]{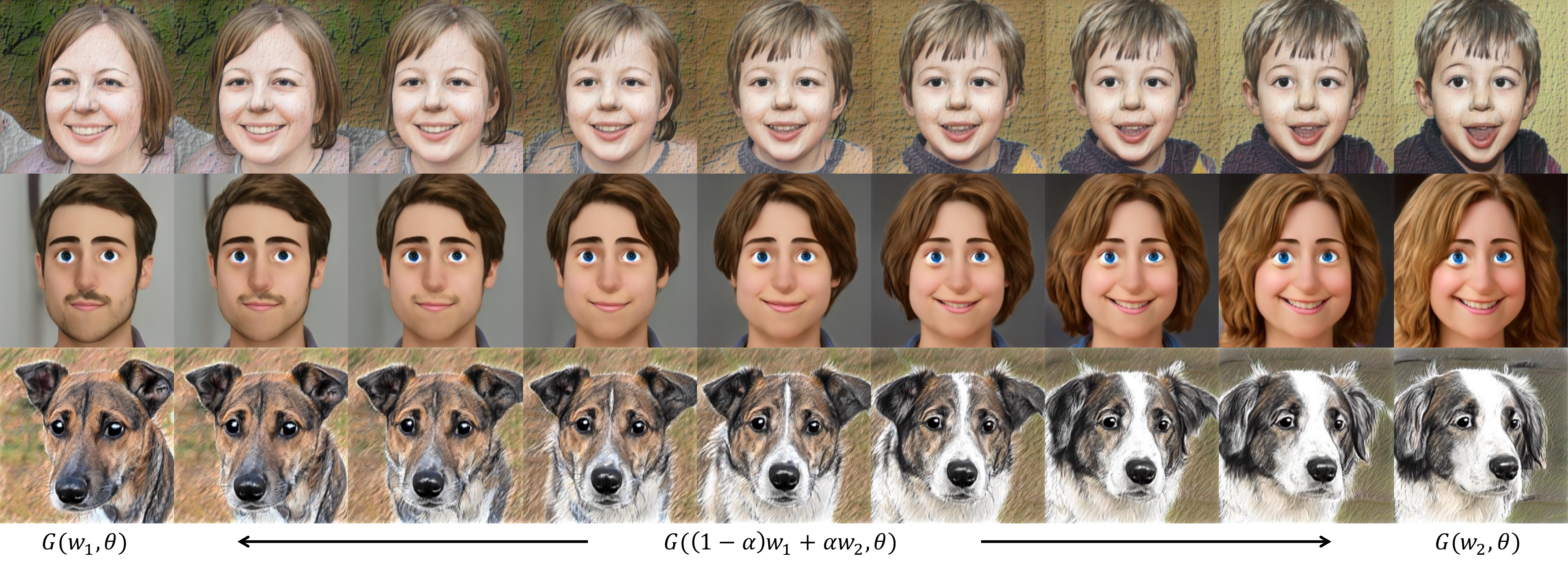}
    \vspace{-2mm}   
  \caption{Visual results of latent space interpolation. The source domain is ``Photo" while the target domains are "Wall painting", "Pixar character" and "Cartoon" from top to bottom. For each row, the left-most column and right-most column are respectively two images synthesized with two different latent codes. The remaining columns refer to images synthesized with interpolated latent codes.}
  \label{fig:sup1}
  \vspace{-2mm}
\end{figure*}
 \begin{figure*}[ht]
  \centering
  \includegraphics[width=0.99\linewidth]{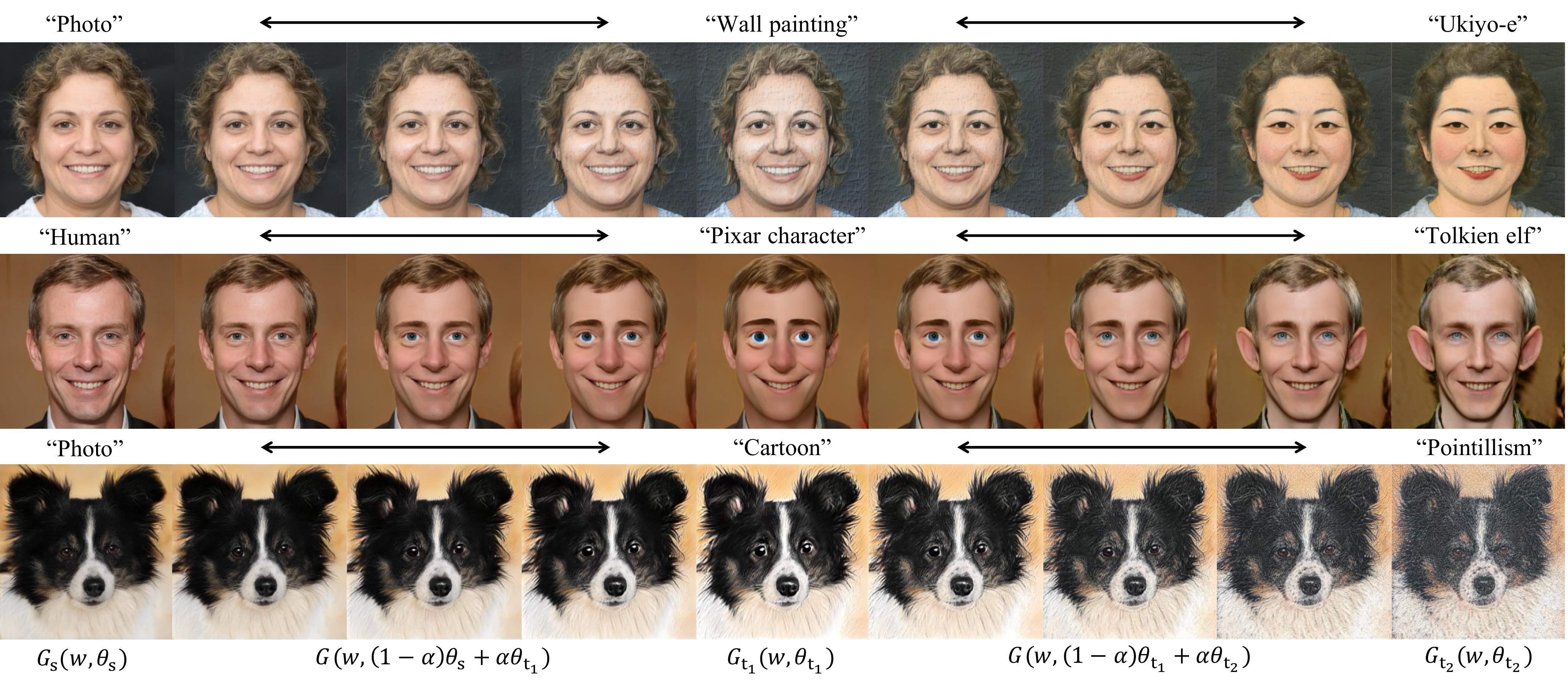}
    \vspace{-2mm}   
  \caption{Visual results of cross-model interpolation. In each row, the left-most image is generated by the source-domain generator. The middle and the right-most images are synthesized by two different target-domain generators. The other images represent cross-model interpolations between two different domains.}
  \label{fig:sup2}
  \vspace{-2mm}
\end{figure*}
 \section{Latent Space Interpolation}

The state-of-the-art generative models \cite{karras2019style,karras2020analyzing,karras2020training,karras2021alias} all have smooth latent spaces for source-domain image generation. We show that the target-domain generators obtained by our method also preserve this superiority.  In \cref{fig:sup1}, each row contains a sequence of images from the same target domain, the left-most column and right-most column are respectively two images $G_{\rm t}(w_1)$ and $G_{\rm t}(w_2)$ synthesized with two different latent codes $w_1$ and $w_2$. For latent space interpolation, an interpolated image is $G_{\rm t}((1-\alpha) w_1 + \alpha w_2)$, where $\alpha \in [0,1]$. For each row, images from left to right correspond to $\alpha$ ranging from 0 to 1. The visual results show that our method has good robustness and generalization ability. The various target-domain spaces obtained by our method are consistently smooth.

\section{Cross-model Interpolation}
Beyond latent space interpolation, we also showcase the model weight smoothness across different domains. In specific, we adopt linear interpolation in weight space for either $G_{\rm s}(\cdot,\theta_{\rm s})$ and  $G_{\rm t}(\cdot,\theta_{\rm t})$ or $G_{\rm t_1}(\cdot,\theta_{\rm t_1})$ and $G_{\rm t_2}(\cdot,\theta_{\rm t_2})$, where $G_{\rm s}(\cdot,\theta_{\rm s})$ denotes the source domain generator, $G_{\rm t_1}(\cdot,\theta_{\rm t_1})$ and $G_{\rm t_2}(\cdot,\theta_{\rm t_2})$ denote two adapted generators of different target domains.
For example, let $\theta_1$, and $\theta_2$ represent the model weights of two generators. Given a latent code $w$, we generate the corresponding image by an interpolated model, $G(w, (1-\alpha) \theta_1+\theta_2)$, where $\alpha \in [0,1]$.
 \cref{fig:sup2} shows that our method has good cross-model interpolation ability, either from a source domain to a target domain or between different target domains.

\section{Geometry Adaptation} 
\begin{figure}[h]
    \begin{center}
\vspace{-2mm}    
\centerline{\includegraphics[width=0.95\columnwidth]{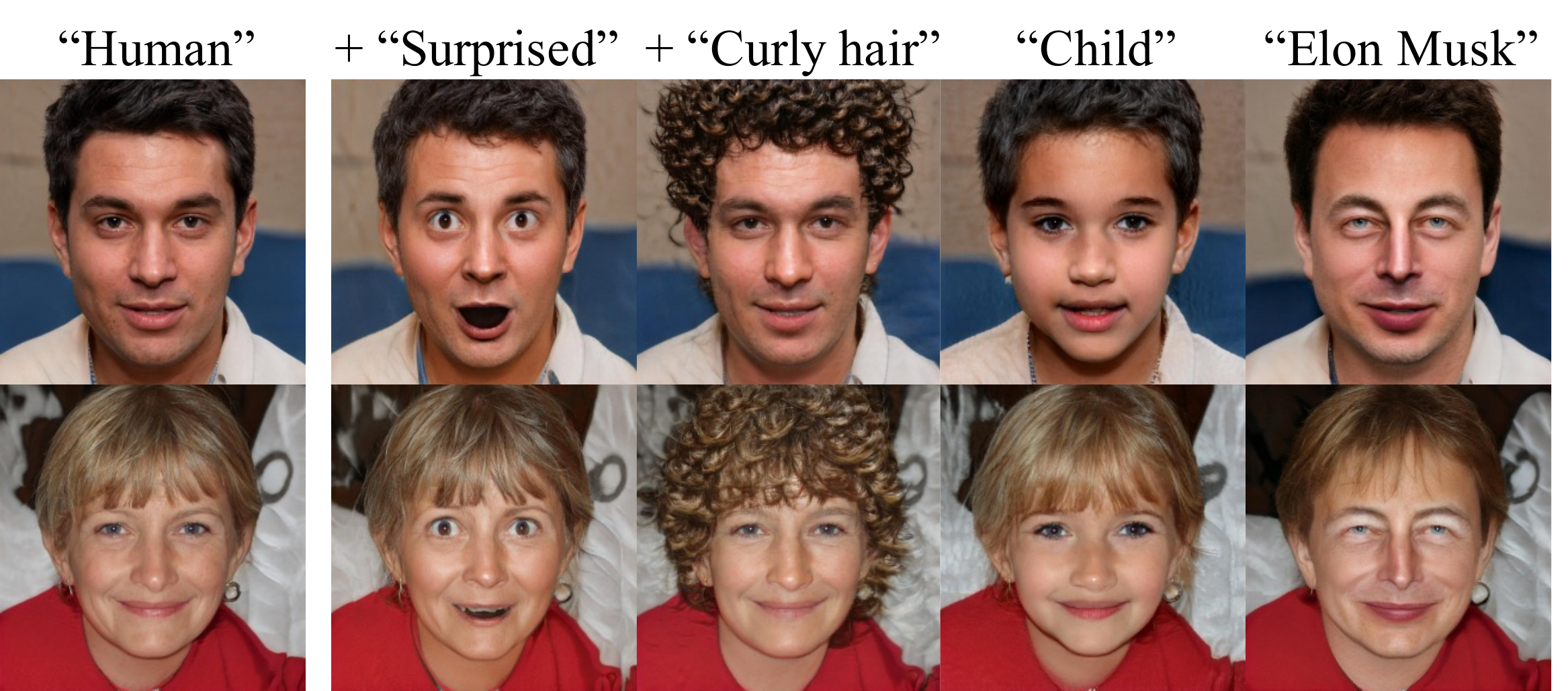}}
    \vspace{-2mm}
    \caption{Geometry adaptation results of IPL.}
    \label{fig:geo}
    \end{center}
    \vspace{-5mm}
\end{figure}

As shown in \cref{fig:geo}, IPL can make diversified geometric edits, such as emotion, haircut, age, and identity like other image manipulation methods.

\newpage

\section{More Well-directed Prompts}

\begin{figure}[h]
  \centering
  \includegraphics[width=0.99\linewidth]{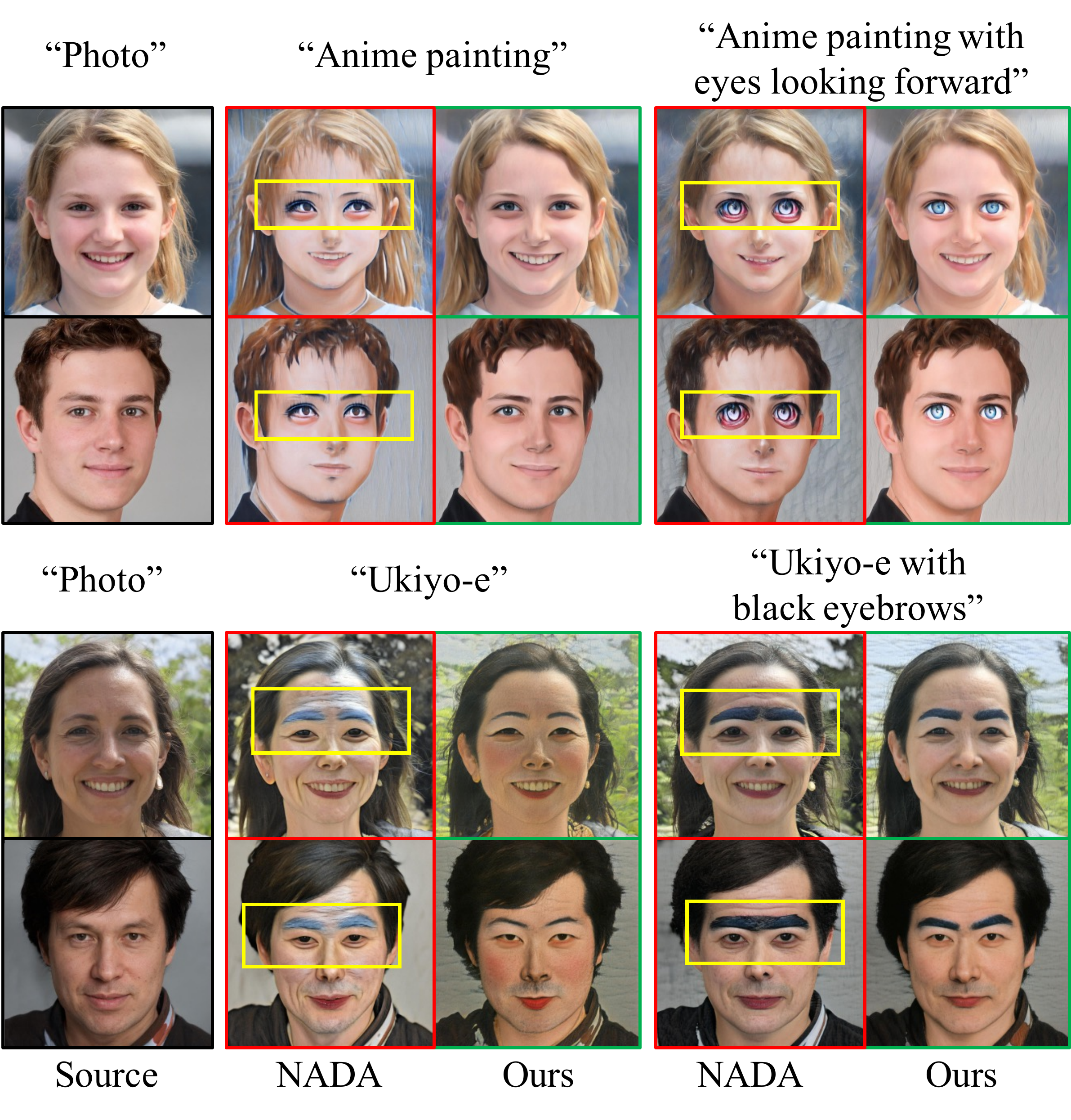}
    \vspace{-2mm}   
  \caption{Image synthesis comparison results with more detailed prompts. The source domain is ``Photo" and the target domains are ``Anime painting" and ``Ukiyo-e". Additional prompts are ``with eyes looking forward" and ``with black eyebrows" for ``Anime painting" and ``Ukiyo-e", respectively. The yellow box areas show the mode collapse patterns of NADA \cite{gal2022stylegan}.}
  \label{fig:sup3}
  \vspace{-4mm}
\end{figure}

A straightforward way to alleviate the mode collapse issue is to manually design a set of well-directed prompts. For example, introduce ``with eyes looking forward" as additional prompts to reduce the squinting eyes issue in “Anime painting”, or use ``with black eyebrows" to solve the blue eyebrows issue in “Ukiyo-e”. In \cref{fig:sup3}, we show that these detailed prompts may lead to other undesired patterns. For ``Anime painting", although the squinting eyes issue can be partly addressed, the generated images of NADA show some similar bleeding eyes patterns. For ``Ukiyo-e", the thick black eyebrows replace the original blue eyebrows for generated results of NADA, but the connecting two eyebrows together is a new undesired pattern. It is worth noting that our IPL is still better than NADA with these additional text prompts and avoids undesired patterns.

\vspace{-2mm} 

 \section{Quantitative Results of Diffusion Models}

 To quantify the performance improvement of Diff-IPL compared to Diff-CLIP \cite{kim2022diffusionclip} and Diff-CLIP+, IS, SCS, ID and SIFID are evaluated. As illustrated in \cref{table:diff}, Diff-IPL performs the best IS, SCS and ID on the two settings, indicating its superiority in the diversity and quality of generated images, together with the structure and identity preservation capability compared to source images. In addition, Diff-IPL achieves the best SIFID score in most cases, showcasing that our method generates the desired target-domain style better. 

 \begin{table}[h]
\begin{center}
\scriptsize
\caption{Quantitative evaluation results of Diff-CLIP \cite{kim2022diffusionclip}, Diff-CLIP+ and Diff-IPL. S$\rightarrow$T, P$\rightarrow$WP and H$\rightarrow$TE denote Source$\rightarrow$Target, Photo$\rightarrow$Wall painting and Human$\rightarrow$Tolkien elf, respectively. The best results are \textbf{bold}.}
  \vspace{-1mm}
\label{table:diff}
\renewcommand{\arraystretch}{1.5}
\setlength{\tabcolsep}{3pt}
\resizebox{\linewidth}{!}{
\begin{tabular}{ccccccccc}
\toprule
\multirow{2}*{S$\rightarrow$T}& \multirow{2}*{Method} &{{\multirow{2}*{IS \cite{salimans2016improved} ($\uparrow$)}}}&{{\multirow{2}*{SCS \cite{xiao2022few} ($\uparrow$)}}}&{{\multirow{2}*{ID \cite{deng2019arcface} ($\uparrow$)}}}&\multicolumn{3}{c}{{SIFID  \cite{shaham2019singan} ($\downarrow$)}}\\
  &&&&&$\rm{R_1}$&$\rm{R_2}$&$\rm{R_3}$\\

\midrule
 \multirow{3}*{P$\rightarrow$WP}&Diff-CLIP&1.696&0.662&0.595&5.493&5.066&5.727\\

 &Diff-CLIP+&2.542&0.611&0.554&2.644&2.099&2.455\\
 &Diff-IPL&\textbf{2.953}&\textbf{0.744}&\textbf{0.785}&\textbf{2.022}&\textbf{1.841}&\textbf{2.004}\\

    \midrule
  \multirow{3}*{H$\rightarrow$TE}&Diff-CLIP&2.055&0.684&0.328&6.091&8.138&7.779\\

 &Diff-CLIP+&2.711&0.627&0.399&\textbf{2.218}&4.283&4.055\\
 &Diff-IPL&\textbf{2.893}&\textbf{0.696}&\textbf{0.709}&2.749&\textbf{3.421}&\textbf{3.696}\\

          \bottomrule
\end{tabular}
}
\end{center}
  \vspace{-7mm}
\end{table}

\section{Diffusion Models versus GANs} We compare the Inception Score results of diffusion models and GANs in \cref{table:vs}. The superiority of Diff-CLIP+ over NADA indicates diffusion models can handle more cases with better base performance. Assisted with IPL, GAN-IPL showcases competitive performance to Diff-CLIP+. Moreover, integrating IPL with Diff-CLIP+ as Diff-IPL also leads to a significant improvement, indicating IPL's compatibility with both GANs and diffusion models.

\begin{table}[h]
\begin{center}
\scriptsize
\caption{Quantitative comparison of GANs and diffusion models. We compare the Inception Score ($\uparrow$) \cite{salimans2016improved} results for Photo$\rightarrow$Wall painting and Human$\rightarrow$Tolkien elf.}
  \vspace{-1mm}
\label{table:vs}
\renewcommand{\arraystretch}{1.5}
\setlength{\tabcolsep}{3pt}
\resizebox{\linewidth}{!}{
\begin{tabular}{ccccc}
\toprule

Source$\rightarrow$Target&NADA (GAN)&GAN-IPL&Diff-CLIP+&Diff-IPL\\
\midrule
 
 Photo$\rightarrow$Wall painting& 2.183 & 2.676 & 2.542 & 2.953 \\

  Human$\rightarrow$Tolkien elf& 2.479 & 2.778 & 2.711 & 2.893\\    
          \bottomrule
\end{tabular}
}
\end{center}
  \vspace{-7mm}
\end{table}

\begin{figure*}[ht]
  \centering
  \includegraphics[width=0.95\linewidth]{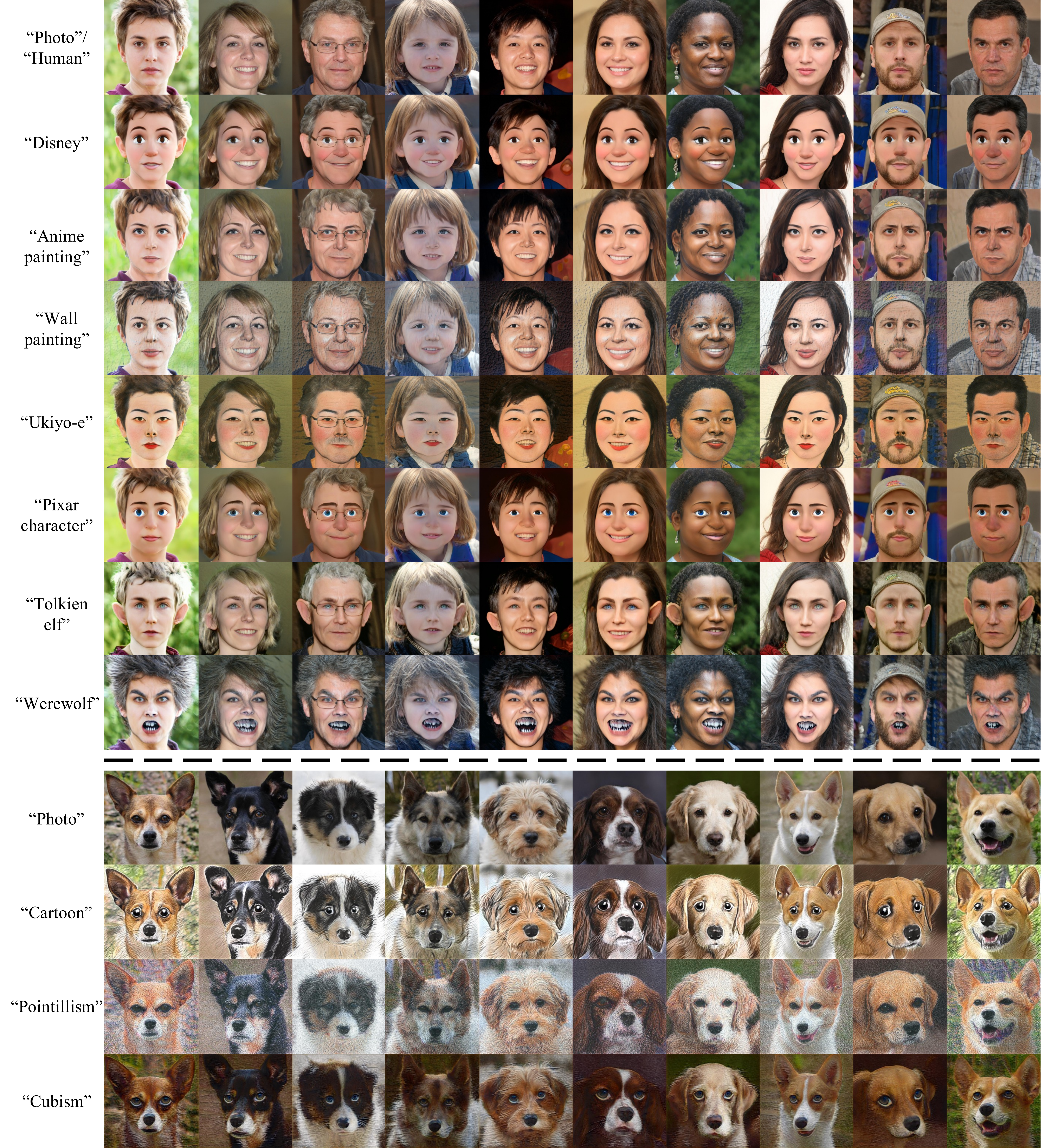}
    \vspace{-1mm}   
  \caption{Additional results of GAN-IPL.}
  \label{fig:sup5}
  \vspace{-4mm}
\end{figure*}

\begin{figure*}[ht]
  \centering
  \includegraphics[width=0.95\linewidth]{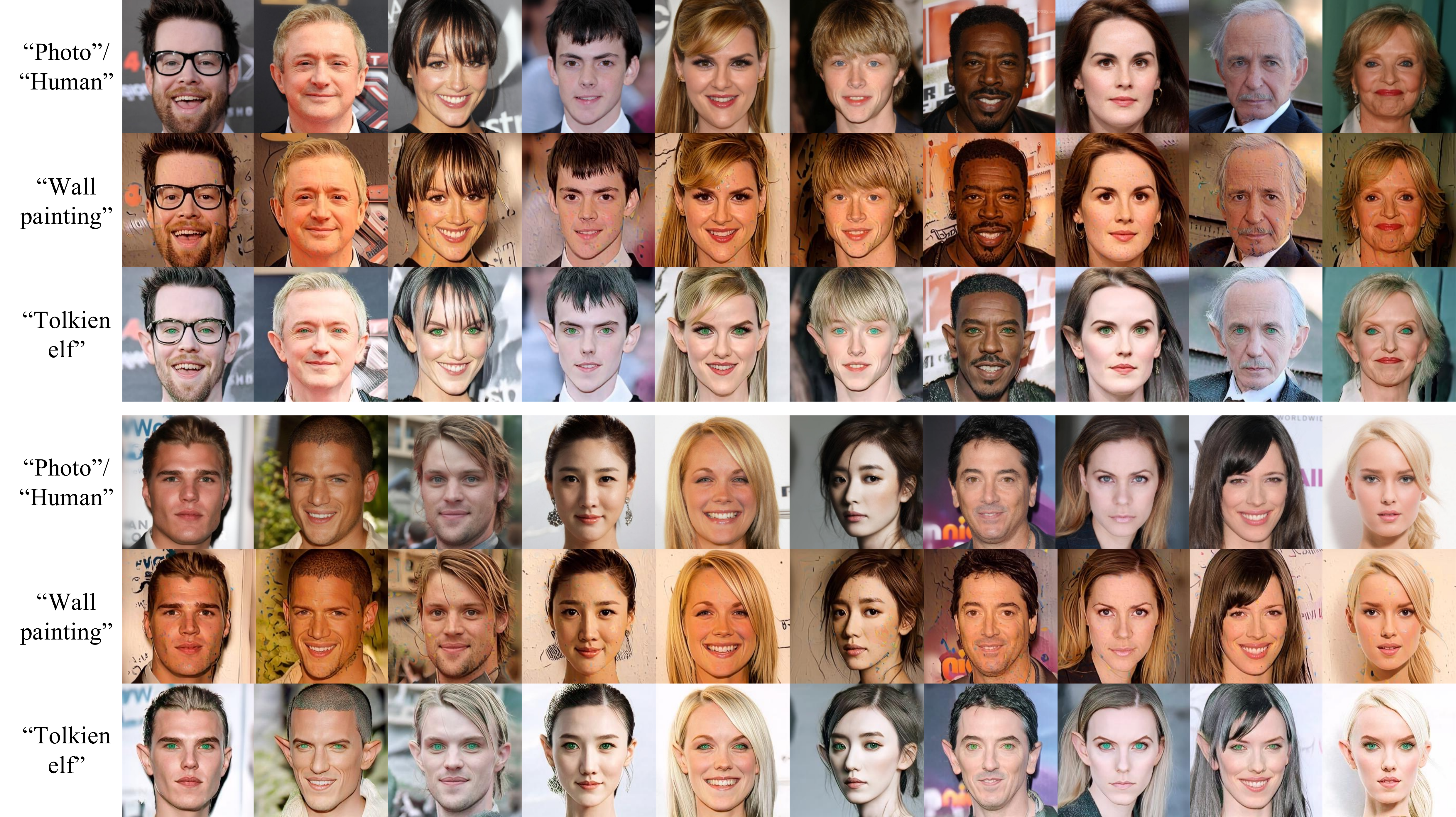} 
  \caption{Additional results of Diff-IPL.}
  \label{fig:sup6}
  \vspace{-2mm}
\end{figure*}
\section{Effect of the Number of Prompts}

 To ensure comparison fairness, we adopted the setting $m=4$ in experiments. In \cref{table:ablmb}, we investigate the effect of $m$ by setting it as 1,2,4,8,16, with the same 300 training iterations. The Inception Score results show that learned prompts (results of different $m$) consistently exceed the manual prompts (NADA). In addition, too small or too large $m$ may lead to insufficient learning and performance degradation. Overall, $m \in [4,8]$ can be optimal. 
 
\vspace{-2mm}
\begin{table}[h]
\begin{center}
\scriptsize
\caption{Quantitative results of different $m$. We evaluate the Inception Score ($\uparrow$) \cite{salimans2016improved} for Photo$\rightarrow$Ukiyo-e.}
  \vspace{-2mm}
\label{table:ablmb}
\renewcommand{\arraystretch}{1.5}
\setlength{\tabcolsep}{3pt}
\resizebox{\linewidth}{!}{
\begin{tabular}{ccccccc}
\toprule

Source$\rightarrow$Target&NADA&$m=1$&$m=2$&$m=4$&$m=8$&$m=16$\\
\midrule
 
 Photo$\rightarrow$Ukiyo-e& 2.205 & 2.757 & 2.943 & 2.974 & 3.047 & 2.651 \\

          \bottomrule
\end{tabular}
}
\end{center}
  \vspace{-7mm}
\end{table}

\section{More Visual Results}
We provide more visual results of GAN-IPL and Diff-IPL across all target domains mentioned in \cref{sec:exp}. In specific, we display additional generative model adaptation results for GAN-IPL in \cref{fig:sup5} and real-world image translation results for Diff-IPL in \cref{fig:sup6}. Although Diff-IPL has a stronger inversion capability for real images (discussed in \cref{sec:inversion}), the visual results of GAN-IPL and Diff-IPL seem to be comparable for general cases. In practice, GAN-IPL is more suitable for applications where plenty of target-domain images are required, since GANs perform a more efficient generative process than diffusion models. While Diff-IPL is more appropriate for applications where the structure and identity of source-domain images need to be precisely preserved in target-domain images.

\begin{figure}[h]
  \centering
  \includegraphics[width=0.99\linewidth]{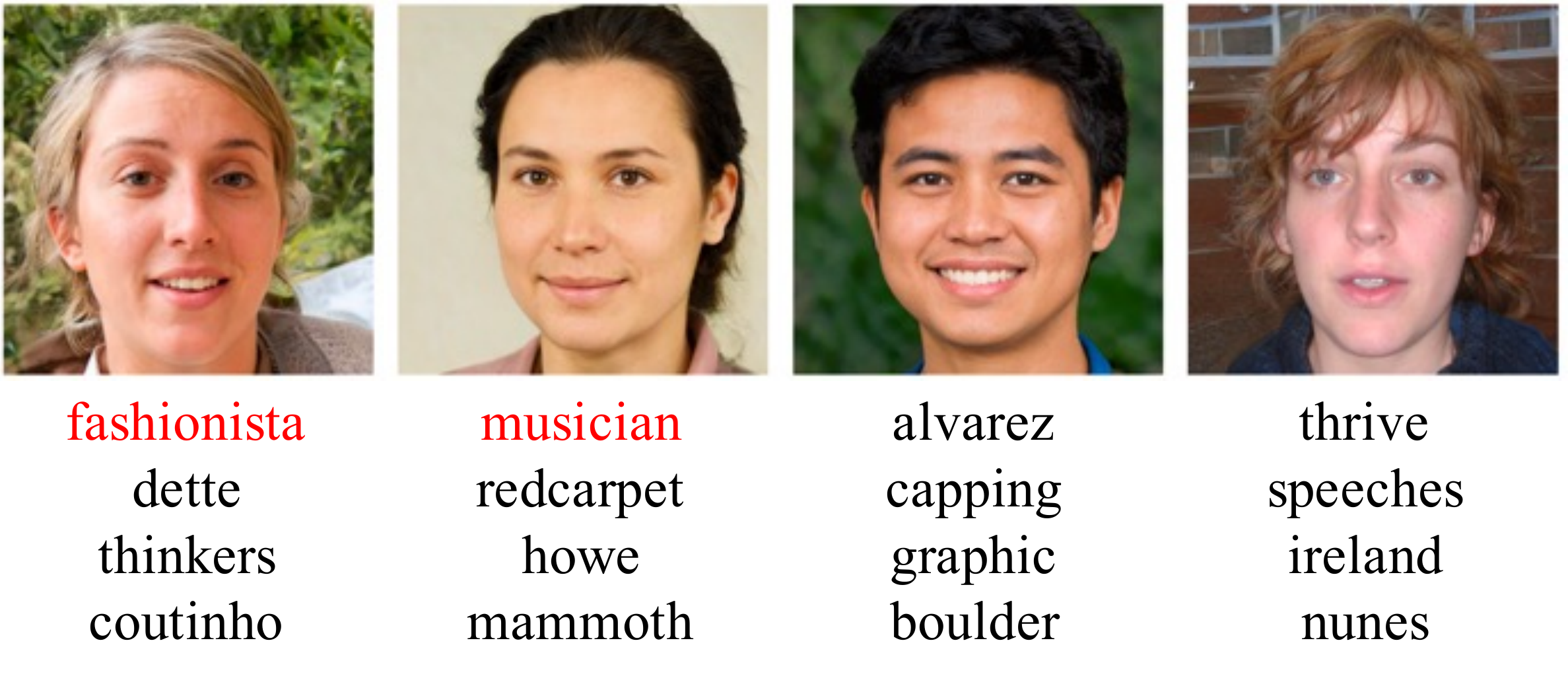}
  \caption{Visualization of the learned prompt vectors. For each image, we present the nearest words of the prompt vectors computed in the word embedding space. \textcolor{red}{Red} words may be somewhat relevant to corresponding images.}
  \label{fig:sup4}
  \vspace{-3mm}
\end{figure}
\section{Prompt Visualization}
Since the prompt vectors are continually optimized, there is no one-to-one correspondence between learned prompt vectors and realistic words. Even so, we try to find some relationships by searching the closest word within the vocabulary for every prompt vector. Following CoOp \cite{zhou2021learning},the Euclidean distance between a prompt vector and the embedding of the a realistic word is computed. We present several cases of these searched image-specific words in \cref{fig:sup4}. Overall, our discovery is similar to the discussion in \cite{zhou2021learning}. A few words are somewhat relevant to their corresponding image, e.g., fashionista, thinkers and musician, while most of the words remain difficult for us to find their connection to images. We conjecture that a source image should contain rich and diverse image-specific semantics. With the limited prompt length, one prompt vector may contain an integration of many different semantics and can not be correctly interpreted with the closest word in the existing vocabulary.

\textbf{Limitation.} Sincerely, the unknown visualization of learned prompt vectors may somewhat limit the interpretability of IPL. We expect that future works could investigate better solutions to effectively decouple and visualize the semantics of a continually optimized prompt vector.

\section{Large Domain Shift.}

\begin{figure}[t]
    \begin{center}
    \centerline{\includegraphics[width=0.99\columnwidth]{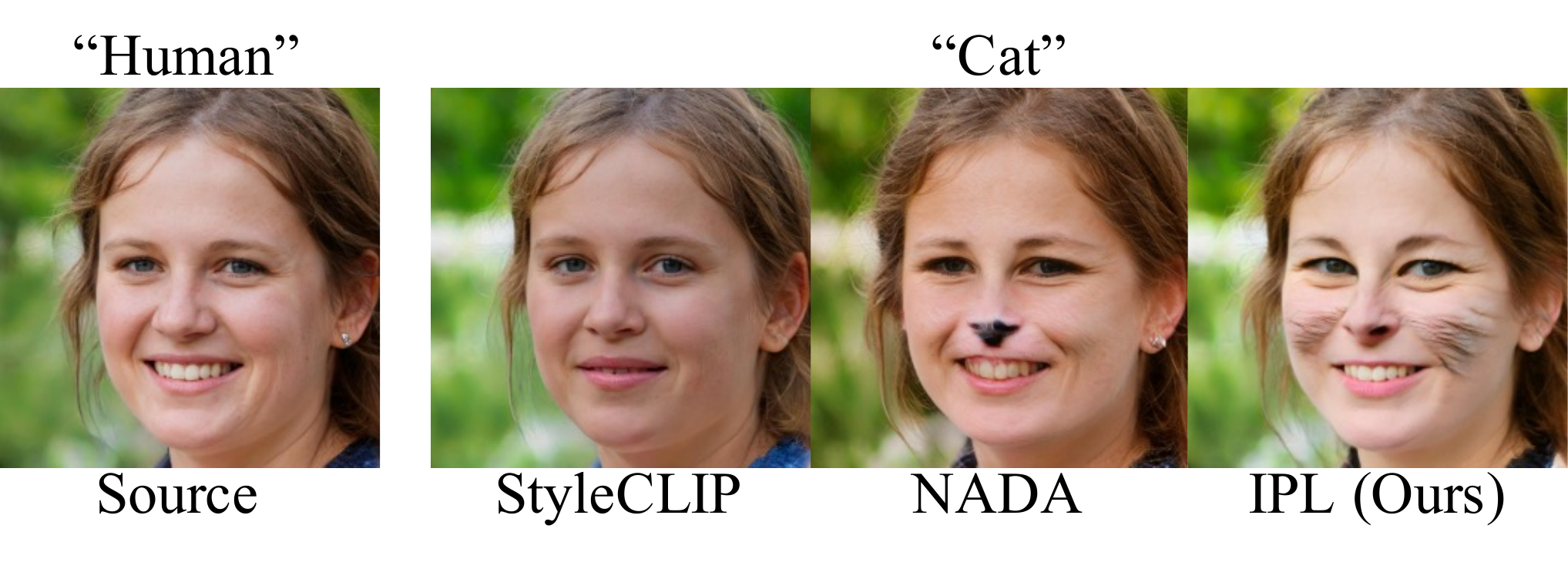}}
    \vspace{-2mm}
    \caption{Image synthesis comparison results with a large domain shift. The source domain is ``Human" and the target domain is ``Cat". We compare IPL with StyleCLIP \cite{patashnik2021styleclip} and NADA \cite{gal2022stylegan}.}
    \label{fig:cat}
    \end{center}
    \vspace{-8mm}
\end{figure}

In general, there is a strong correlation between source and target domains in domain adaptation tasks. As demonstrated in \cref{fig:cat}, generator adaptation with a large domain shift (e.g., from ``Human'' to ``Cat'') is challenging for all existing zero-shot generators and requires future investigation. However, we can observe that IPL could present more cat-like whiskers and eyes, compared with other zero-shot competitors, i.e., StyleCLIP \cite{patashnik2021styleclip} and NADA \cite{gal2022stylegan}.

\section{Social Impact}
IPL may contribute to artistic image synthesis applications in social media industries. It may also assist the other computer vision tasks (e.g., recognition and detection) as a data augmentation technique. However, the ability of IPL to synthesize fake images from real-world images may bring some ethical problems, which must be treated carefully.
\end{document}